\RequirePackage{fix-cm}
\documentclass[smallextended]{svjour3}       
\smartqed  
\usepackage[colorlinks,linkcolor=red,anchorcolor=green,citecolor=blue]{hyperref}
\usepackage{graphicx}
\usepackage{amsfonts,amssymb}
\usepackage{amsmath}
\usepackage{enumerate}
\usepackage[ruled,vlined]{algorithm2e}
\usepackage[labelformat=simple]{subcaption}

\usepackage{verbatim}

\usepackage{xcolor}
\usepackage{lineno}

\begin{document}

\title{A deep neural network framework for dynamic multi-valued mapping estimation and its applications \thanks{This work is supported by HKRGC  GRF  (Project  ID: 14305919).}
}

\author{Geng LI \and
        Di QIU \and
        Lok Ming LUI
}


\institute{Geng LI \at
            The Chinese University of Hong Kong \\
              \email{gli@math.cuhk.edu.hk}
           \and
           Qiu DI \at
           Google Research \\
           \email{sylvesterqiu@gmail.com}
           \and
           Lok Ming LUI \at
           The Chinese University of Hong Kong \\
           \email{lmlui@math.cuhk.edu.hk}
}

\date{Received: date / Accepted: date}

\maketitle

\begin{abstract}
This paper addresses the problem of modeling and estimating dynamic multi-valued mappings. While most mathematical models provide a unique solution for a given input, real-world applications often lack deterministic solutions. In such scenarios, estimating dynamic multi-valued mappings is necessary to suggest different reasonable solutions for each input. This paper introduces a deep neural network framework incorporating a generative network and a classification component. The objective is to model the dynamic multi-valued mapping between the input and output by providing a reliable uncertainty measurement. Generating multiple solutions for a given input involves utilizing a discrete codebook comprising finite variables. These variables are fed into a generative network along with the input, producing various output possibilities. The discreteness of the codebook enables efficient estimation of the output's conditional probability distribution for any given input using a classifier. By jointly optimizing the discrete codebook and its uncertainty estimation during training using a specially designed loss function, a highly accurate approximation is achieved. The effectiveness of our proposed framework is demonstrated through its application to various imaging problems, using both synthetic and real imaging data. Experimental results show that our framework accurately estimates the dynamic multi-valued mapping with uncertainty estimation.
\end{abstract}

\section{Introduction}\label{sec_introduction}
\begin{figure*}[t]
	\centering
	\includegraphics[width=5.5in]{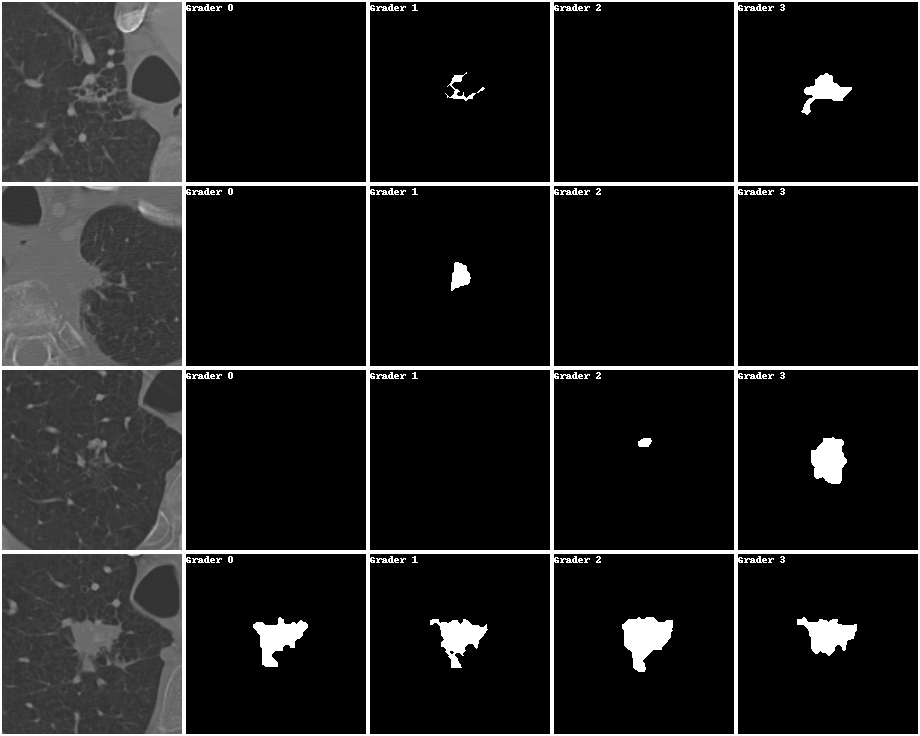}
	\caption{Some samples of lung CT scans. The first column is lung CT scans and the left column is labels from four experts. Experts provide different annotations for each CT scan, resulting in varying numbers and probabilities of outputs. }
	\label{sample} 
\end{figure*}

Uncertainty is a significant challenge that arises when making predictions for various tasks, including pose estimation, action prediction, and clinical diagnosis. In many cases, obtaining an accurate and definitive solution is difficult due to missing information or noise. There can be multiple possible interpretations or solutions. Moreover, in complex real-world scenarios, the relationship between inputs and solutions becomes even more intricate. Different inputs may correspond to different numbers of potential solutions. For example, in natural language processing, a sentence may have two different meanings if contextual information is lacking, while another sentence may have three meanings. Similarly, in diagnosing lung lesions, different medical professionals may provide varying diagnoses for a patient's CT scan showing potential lung damage. This variation in diagnosis can be attributed to incomplete information and the inherent uncertainties in medical imaging. If we consider all the diagnoses provided by different medical professionals as potential solutions, each scan can have a varying number of potential diagnoses (see Fig \ref{sample}). Therefore, there is a need to develop an effective framework that can address the challenge of uncertainty estimation in real-life scenarios.

Mathematically, the aforementioned problems can be described as follows: Let $X$ and $Y$ represent the input space and output space, respectively. We are provided with a collection of paired datasets $\{{(x_i,y_i^{k})}_{k=1}^{\mathcal{N}_i}\}_{i=1}^\mathcal{T}$, where $y_i^k \in Y$ is a possible output corresponding to $x_i \in X$. These paired datasets consist of unorganized pairs, allowing for flexibility as each collection $\{y_i^{k}\}_{k=1}^{\mathcal{N}_i}$ may contain repetitions. Our objective is to find a suitable mapping $f$ that fits this dataset. However, the required mapping for this type of dataset is non-standard. For each $x\in X$, $f(x) = \{y_j \in Y\}_{j=1}^{\mathcal{N}_{x}}$, where $\mathcal{N}_{x} \in \mathbb{N}$ depends on $x$ and $y_j$'s are distinct for different $j$. In other words, the number of plausible outputs associated different $x\in X$ may vary, and hence $\mathcal{N}_{x}$ is dynamic depending on $x\in X$. We refer to this type of mapping as a {\it dynamic multi-valued mapping (DMM)}. To find an optimal DMM that fits the given dataset, we need to solve the following mapping problem:
\begin{equation}
f = \mathop{\arg\min}\limits_{g: X \rightarrow \mathcal{P}^{f}(Y)} \mathcal{L}(g)
\end{equation}

\noindent Here, $\mathcal{L}$ is a suitable loss function, such as the $L^2$ fidelity data loss, that depends on the datasets and applications. In addition to fitting the dataset, it is also desirable to estimate the probability of each plausible output. The likelihood is dependent on the occurrence of an output in the input dataset. However, an immediate challenge we face is how to mathematically model a DMM. Directly implementing such a mapping can be mathematically challenging, particularly when dealing with the diversity of $\mathcal{N}_{x}$. Motivated by this challenge, we are interested in developing a numerical framework to solve the dynamic multivariate mapping problem described above.

In many scenarios, the number of possible outputs $\mathcal{N}_{x}$ is bounded by a fixed number $N \in \mathbb{N}$. In other words, for each input $x \in X$, the number of potential outputs is always less than or equal to $N$. In this case, a DMM can be viewed as a one-to-$N$ mapping with uncertainty estimation. To measure the likelihood of each plausible output, we introduce probability measure for each output while disregarding those with zero probabilities. This approach allows us to obtain a sequence of results with a dynamic number of elements $\mathcal{N}_{x}$, which depends on the input $x$. Although the size of $\mathcal{N}_{x}$ is constrained by a predefined value $N$, we can effectively handle most real-world problems by selecting an appropriate value for $N$. To model a one-to-$N$ mapping with uncertainty estimation, we can utilize a dictionary $C = \{c_1, c_2, ..., c_N\}$, also known as a {\it codebook}, which consists of a collection of finite variables. Thus, a DMM can be described by two bivariate functions: $f : X \times C \rightarrow Y^N$ and $p : C \times X \rightarrow [0, 1]^N$. For each $j = 1, 2, ..., N$, $f(x, c_j)$ represents a potential output associated with input $x$, while $p(c_j | x)$ represents the probability of generating such an output $f(x, c_j)$. Outputs with a probability of $0$ are discarded. This formulation allows us to effectively represent a DMM as two bivariate functions. In this work, our numerical framework to solve the DMM problem is based on this construction. For ease of computation, the bivariate functions $f$ and $p$ are parameterized by deep neural networks. Different parameterizations of the deep neural networks result in different $f$ and $p$ functions. Consequently, the optimal DMM can be obtained by optimizing the parameters of the deep neural networks.

Numerous studies have explored the use of auto-encoders to model multi-valued mappings by sampling and decoding latent codes from the latent space. Conditional generation has shown promise in producing multiple outcomes for various tasks. Noteworthy examples include the works of Zhu et al. \cite{zhu2017toward} and Huang et al. \cite{huang2018multimodal}, which focus on the task of multi-modal image-to-image translation. Additionally, Zheng et al. \cite{zheng2019pluralistic} introduced a specialized approach for image inpainting. These methods demonstrate the capability to generate a variety of plausible outputs. However, these methods often lack uncertainty estimation for each plausible output, making them inadequate for solving the DMM problem. One notable exception is the Probabilistic U-Net proposed by Kohl et al. \cite{kohl2018probabilistic}. Built upon the conditional VAE framework \cite{sohn2015learning}, the Probabilistic U-Net allows for quantitative performance evaluation thanks to its application and associated datasets. Moreover, the Probabilistic U-Net outperforms many other methods in calibrated uncertainty estimation, including the Image2Image VAE of \cite{zhu2017toward}. However, it inherits the Gaussian latent representation from CVAE \cite{sohn2015learning}, which leads to the drawback of posterior collapse, resulting in wrong outputs in the generated samples. In our previous conference paper \cite{qiu2021modal}, we proposed a preliminary method to address the challenges faced by auto-encoders in modeling multi-valued mappings. By utilizing a discrete representation space to approximate the multi-modal distribution of the output space, our preliminary method aimed to overcome the issue of posterior collapse and provide rough estimates of the conditional probability associated with each output. However, further analysis revealed several drawbacks. Firstly, the preliminary method exhibited repetitions in outputs, where the same output corresponded to different modes, rendering it unsuitable for representing a DMM. Secondly, the performance of the preliminary method suffered when dealing with imbalanced datasets, as the accuracy of the uncertainty estimation associated with different modes was often inaccurate. Lastly, the output results of the preliminary method were found to be inaccurate when dealing with imbalanced and unorganized dataset. These drawbacks make it challenging to use the preliminary method to represent a DMM.

In this work, building upon our previous preliminary model \cite{qiu2021modal}, we develop a deep neural network framework that is capable of addressing the DMM problem. The bivariate functions $f$ and $p$ are parameterized by a deep generative network $G_{\theta}$ and a classification network $P_{\theta}$, respectively. The generative network $G_{\theta}$ is responsible for generating multiple results as a one-to-$N$ mapping $f$, while the classification network $P_{\theta}$ predicts the probabilities associated with these multiple results. The multiple outcomes of a given input $x$ are associated with a codebook, which is a set of discrete variables ${c_1, c_2, ..., c_N}$. Specifically, we utilize the generative network $G_{\theta}$ to generate multiple results $G_{\theta}(x, c_j)$ based on the discrete variables $c_j$. Simultaneously, our classification network $P_{\theta}$ predicts the probability $P_{\theta}(c_j, x)$ for each variable $c_j$, which serves as the probability for each outcome $G_{\theta}(x, c_j)$. There are two key challenges that need to be addressed in the proposed framework. Firstly, to ensure that the deep neural network framework accurately represents a DMM, it is essential that the outputs $G(x,c_j)$ of the generative network differ for different code values $c_j$. In other words, for each input $x\in X$, the mapping $G_{\theta}(x,\cdot)$ must be injective, guaranteeing the generation of a unique output for each code and preventing duplication. Additionally, another significant challenge is to enrich the codebook with essential information, allowing the generative network to produce diverse and accurate results, while also accurately predicting the probabilities associated with each outcome. To overcome these challenges, we propose a specialized loss function that incorporates the covariance loss and the ETF cross-entropy loss. This loss function enables the generation of diverse, plausible, and unique outputs for each code, as well as giving an accurate uncertainty estimation. Consequently, our framework effectively parameterizes a DMM with a dynamic range of plausible outputs for each input. Even when dealing with imbalanced data, our proposed framework can identify an optimal DMM that fits the given dataset. We evaluate the performance of our proposed framework on various imaging problems, including both synthetic and real images. Experimental results demonstrate the effectiveness of our model in solving dynamic multi-valued mapping problems across a range of imaging applications.


The rest of the paper is organized as follows. The section \ref{sec_contribution} outlines the primary contributions of this work. We introduce our general framework and model details in \ref{sec_model}. In \ref{sec_exp}, real and synthetic datasets are used in model experiments, and some details are introduced. Finally, we conclude the paper in \ref{sec_conclusion}. 

\section{Contribution}\label{sec_contribution}
The main contributions of this paper are listed as follows.

\begin{enumerate}
\item We propose a notion of dynamic multi-valued mapping (DMM) and formulate a general optimization problem for DMM to address the challenge of computing multiple plausible solutions with uncertainty estimation in real-life scenarios.
\item By considering a DMM by a $1$-to-$N$ mapping with uncertainty estimation, a DMM is represented by two bi-variate functions, which are parameterized by deep neural networks. A specialized loss function based on the covariance loss and the incorporation of the ETF cross-entropy loss is proposed to train the deep neural networks. This enables the framework to be capable of representing a DMM, with dynamic number of plausible outputs.
\item We apply the proposed framework to various practical imaging problems, demonstrating its efficacy and effectiveness in solving real-world challenges.
\end{enumerate}

\section{Proposed model}\label{sec_model}
In this section, we will describe our proposed deep neural network framework for computing multiple plausible solutions with uncertainty estimation in real-life scenarios.

\subsection{Mathematical formulation of dynamic multi-valued mapping problem}

In this subsection, we will first provide a mathematical formulation of our proposed problem. Our objective is to develop a framework that suggests multiple plausible outputs and estimates their likelihood based on a given dataset. Let $X$ and $Y$ represent the input space and output space, respectively. Suppose we are given a collection of paired datasets $\mathcal{D} = \{\{(x_i,y_i^k)\}_{k=1}^{\mathcal{N}_i}\}_{i=1}^{\mathcal{T}}$. Here, $y_i^k \in Y$ represents a plausible solution associated with the input $x_i\in X$.  In practice, the datasets $\mathcal{D}$ can be imbalanced and unorganized. For each $i$, $y_i^k$ may be repeated for different values of $k$. For example, when labeling the lesion position for ambiguous medical images, different medical experts may provide the same predictions. Consequently, this results in duplicated outputs within $\mathcal{D}$. On the other hand, quantifying the likelihood of each suggested output can be challenging. The extent of repetition in $\mathcal{D}$ can provide valuable information for estimating the probability associated with each plausible output.

In this context, our proposed problem can be mathematically formulated as finding an appropriate mapping $f$ to fit the given dataset $\mathcal{D}$. The main challenge arises from the fact that each input $x\in X$ can be associated with multiple plausible outputs. Furthermore, the number $\mathcal{N}_{x}$ of plausible outputs corresponding to different inputs $x\in X$ can vary, making it dynamic and dependent on $x$. Traditional mappings between $X$ and $Y$ are inadequate for fitting $\mathcal{D}$. To address this unique dataset, we introduce the concept of a {\it dynamic multi-valued mapping (DMM)} defined as follows:

\begin{definition}
Let $X$ and $Y$ be two metric spaces. A {\it dynamic multi-valued mapping (DMM)} between $X$ and $Y$ is a mapping $f: X\to \mathcal{P}^f(Y) \subset \mathcal{P}(Y)$, where $\mathcal{P}(Y)$ denotes the power set of $Y$ and
$$\mathcal{P}^f(Y) : = \{ \mathcal{S}\in \mathcal{P}(Y): \mathcal{S} \text{ is non-empty and finite}\}.$$
\end{definition}

In other words, for each $x\in X$, a DMM maps $x$ to a non-empty and finite subset of $Y$. We assume that the number of plausible outputs for each input $x\in X$ is finite, and that each input $x\in X$ must be associated with at least one plausible output. These assumptions generally hold true in most imaging problems.

Our objective is to find an optimal DMM $f$ that accurately represents the dataset $\mathcal{D}$. Specifically, for each input $x_i$, we aim to find an optimal $f$ such that $f(x_i)$ is a subset containing all plausible outputs $y_i^1$, $y_i^2$,..., $y_i^{\mathcal{N}_i}$. It is important to note that for each $i$, the same plausible output $y_i^k$ may appear multiple times for different values of $k$. Consequently, the cardinality $|f(x_i)|$ of the subset is always less than or equal to $\mathcal{N}_i$. Additionally, it is desirable to estimate the likelihood of each plausible output by considering the extent of repetition of each output in $\mathcal{D}$. Next, we will discuss how we can mathematically formulate this problem.

For this purpose, the first task is to mathematically model a DMM. In most real-world scenarios, the number of plausible output $\mathcal{N}_{x}$ can be bounded by a fixed number $N\in \mathbb{N}$ for all $x$. That means the number of possible outputs for each input is at most $N$. In this case, a DMM can be represented by a one-to-$N$ mapping $f: X\to Y^N$ or $f(x) = (y_1,y_2,...,y_N)$ for $\forall x\in X$, together with an uncertainty estimation $p(j|x)$. $p(j|x)$ measures the probaility or likelihood of the solution $y_j$. If $p(j|x) = 0$, the output $y_j$ is discarded and we can simply set $y_j = 0$. The probability measure $p$ helps us to model the dynamic nature of $\mathcal{N}_{x}$. More specifically, 
\[
\mathcal{N}_{x} = N - |\{j: p({j} | x) = 0\}|,
\]
\noindent where $|\cdot|$ is the cardinality of a set. Also, we require that $y_j\neq y_k$ if $j\neq k$, $p(j|x)\neq 0$ and $p(k|x)\neq 0$. This requirement is necessary to ensure that $(f,p)$ can effectively represent a DMM.

Under this setup, we can formulate our problem of fitting $\mathcal{D} = \{\{(x_i,y_i^{k})\}_{k=1}^{\mathcal{N}_i}\}_{i=1}^{\mathcal{T}}$ as an optimization problem over the space of DMMs that minimizes:
\begin{equation}
\begin{aligned}
E_1(f,p) = \frac{1}{\mathcal{T}}\sum^{\mathcal{T}}_{i=1} \frac{1}{\mathcal{N}_i}&\sum^{\mathcal{N}_i}_{k=1} \log \left(\frac{1}{p(s_{i}^{k} | x_i)^{\alpha}}d(f(x_i)_{s_{i}^{k}}, \ y_{i}^{k}) \right) \\
\text{such that} \quad & s_{i}^{k} = \mathop{\arg\min}\limits_{s=1,...,N} d'(f(x_i)_{s}, \ y_{i}^{k}),\\
&\sum^{N}_{j=1}p(j|x_i) = 1 , \  \forall \ x_i \in  X.
\end{aligned}    
\label{general_loss}
\end{equation}
where $f(x)_{s}$ is the $s$-th item of $f(x)$, $d(\cdot, \cdot)$ is the data fitting term, and $d'(\cdot, \cdot)$ is the distance functions for index choice. $\alpha>0$ is a fixed parameter. The objective $E_1(f,p)$ aims to encourage the suggested solutions $f(x_i)$ to closely match the given dataset $\mathcal{D}$ by minimizing the discrepancy between the mapped outputs and the true values. Moreover, minimizing $E_1(f,p)$ promotes larger values of $p(s_i^k|x_i)$ when the mapped outputs $f(x_i)_{s_i^k}$ appear more frequently in the paired dataset $\{x_i,y_i^k\}_{k=1}^{\mathcal{N}_i}$. This allows $p$ to capture the repetitive patterns in the dataset.

To ensure the capability of $f$ to represent a DMM, we impose a constraint on $f(x)$ that for each $x\in X$, such that $(f(x))_j \neq (f(x))_k$ if $j\neq k$, and both $p(j|x) \neq 0$ and $p(k|x) \neq 0$. In order to effectively enforce this property, we leverage the concept of a codebook in formulating $f$ and $p$. Let $C = \{c_1, c_2,...,c_N\}$ be a codebook, where each $c_j\in \mathbb{R}^m$. We can now express $f$ and $p$ as bivariate functions: 
\[
\begin{aligned}
& f: X\times C \to Y^N \quad \text{where} \quad f(x,c_j) = y_j, \\
& p: C\times X \to [0,1] \quad \text{where} \quad p(c_j,x) = p(j|x).
\end{aligned}
\]
By formulating $f$ and $p$ in this manner, we can optimize the codebook to control the properties of $f$ and $p$, ensuring their suitability for representing a DMM. Our optimization can now be rewritten as finding two bivariate functions $f: X\times C\to Y^N$ and $p: C\times X\to [0,1]$ by minimizing:

\begin{equation}
\begin{aligned}
E_2(f, p) = \frac{1}{\mathcal{T}}\sum^{\mathcal{T}}_{i=1} \frac{1}{\mathcal{N}_i}&\sum^{\mathcal{N}_i}_{k=1} \log \left(\frac{1}{p(s_{i}^{k}, x_i)^{\alpha}} d(f(x_i, c_{s_{i}^{k}}), \ y_{i}^{k}) \right) \\
\text{such that} \quad & s_{i}^{k} = \mathop{\arg\min}\limits_{s=1,...,N} d'(f(x_i, c_{s}), \ y_{i}^{k}),\\
&\sum^{N}_{j=1}p(j,x_i) = 1 , \  \forall \ x_i \in  X.
\end{aligned}    
\label{rewritten_loss}
\end{equation}

Note that obtaining the code index from the codebook by traversing all the output results can introduce a significant computational burden, especially when dealing with complex output spaces. Additionally, the choice of distance function, denoted as $d'(\cdot, \cdot)$, defined on the output space can greatly impact the performance of our model. If the structure of the output space is excessively complex or overly simplistic, it may lead to suboptimal results when using certain distance functions. To address these challenges, we introduce the concept of cluster mapping, denoted as $z: (X, Y) \rightarrow \mathbb{R}^m$. This mapping allows us to bypass the need for traversing the entire output space and instead focus on finding the most suitable distance metric within the discrete codebook, leading to improved efficiency and effectiveness. The problem of obtaining the index of code can then be formulated as follows:
\begin{equation}\label{closestcode}
s_{i}^{k} = \mathop{\arg\min}\limits_{s=1,...,N} ||z(x_i, y_{i}^{k})- \ c_{s} ||^2
\end{equation}

Additionally, the codebook can also be simultaneously optimized with a suitable regularization to obtain the best collection of codes to capture multiple outputs.  The final optimization problem can now be written as finding optimal $f$, $p$, and $z$, which minimizes:
\begin{equation}
\begin{aligned}
E_3 ({f, p, z, C}) = \frac{1}{\mathcal{T}}\sum^{\mathcal{T}}_{i=1} \frac{1}{\mathcal{N}_i}&\sum^{\mathcal{N}_i}_{k=1} \log \left(\frac{1}{p(c_{s_{i}^{k}}|x_i)^{\alpha}}d(f(x_i, c_{s_{i}^{k}}), \ y_{i}^{k}) \right) \\
& + \beta d''(z(x_i, y_{i}^{k}), c_{s_{i}^{k}}) + \gamma d'''(C)\\
\text{such that} \quad & c_{s_{i}^{k}} = \mathop{\arg\min}\limits_{c \in C} || z(x_i, y_{i}^{k})- \ c ||^2,\\
&\sum^{N}_{j=1}p(c_j|x_i) = 1 , \  \forall \ x_i \in  X.
\end{aligned}    
\label{final_loss}
\end{equation}
where $\beta, \gamma > 0$ are weight parameters. $d''(\cdot, \cdot)$ is a regularization term to restrict $z$. $d'''$ is the regularization term to control the property of $C$.

The primary challenge lies in effectively modeling $f$, $p$, and $z$ in a way that enables simultaneous optimization of the multi-valued mapping and uncertainty estimation. To address this challenge, we adopt a strategy of formulating the problem using a deep neural network, whereby the optimization problem is solved through training the network parameters. The specific details of this approach will be elaborated upon in the subsequent subsection.

\subsection{Deep neural network framework for DMM problem}
As outlined in the previous subsection, the problem at hand can be formulated as an optimization problem involving three bi-variate mappings: $f$, $p$, and $z$. In this work, we propose to parameterize these mappings using deep neural networks. In this subsection, we will provide a comprehensive explanation of our proposed deep neural network framework for the dynamic multi-valued mapping (DMM) problem.

\subsubsection{Overall Network structure}
\begin{figure}[t]
	\centering
	\includegraphics[width=\linewidth]{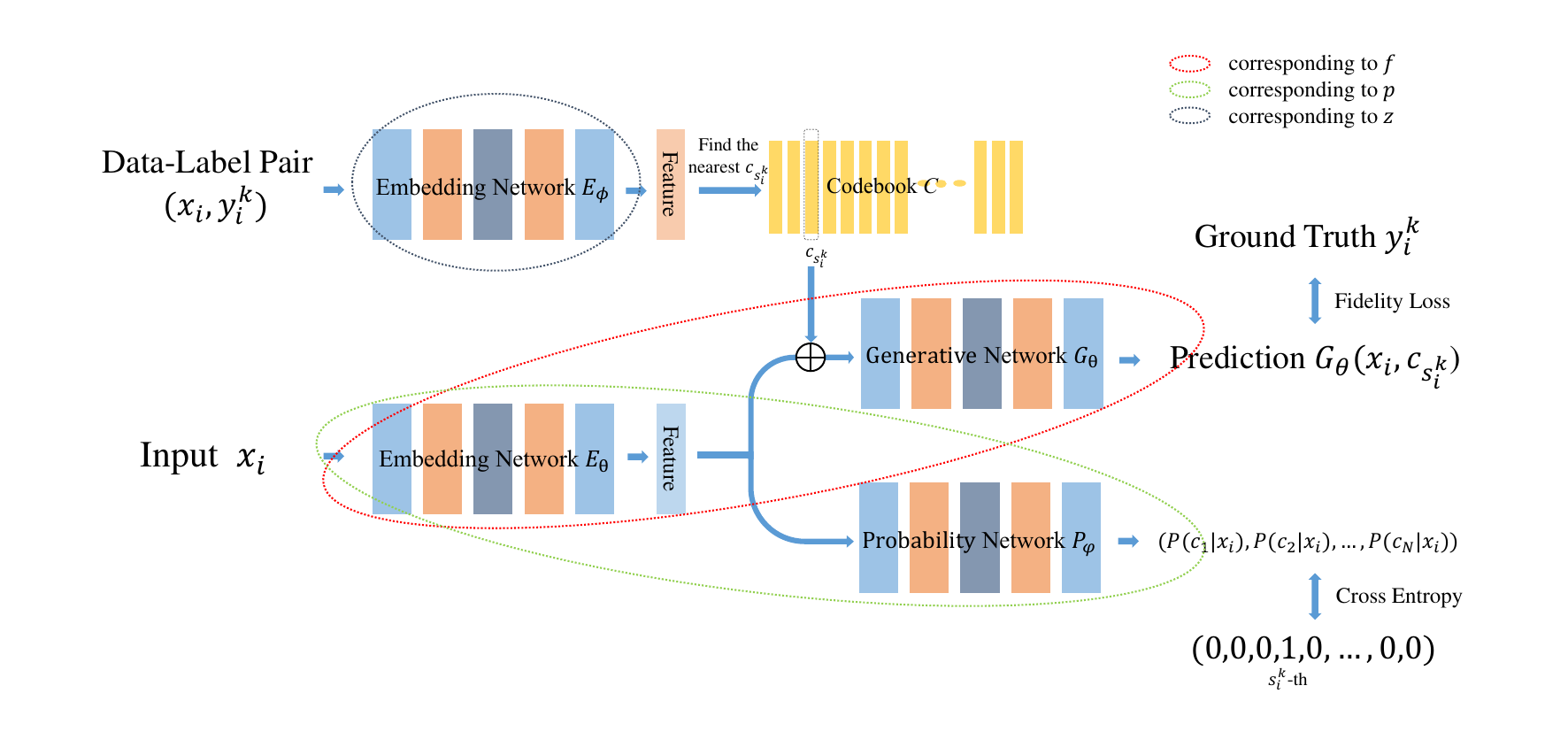}
	\caption{The architecture of our model in the training process.}
        \label{dmm_structure}
\end{figure}

\begin{figure}[t]
	\centering
	\includegraphics[width=\linewidth]{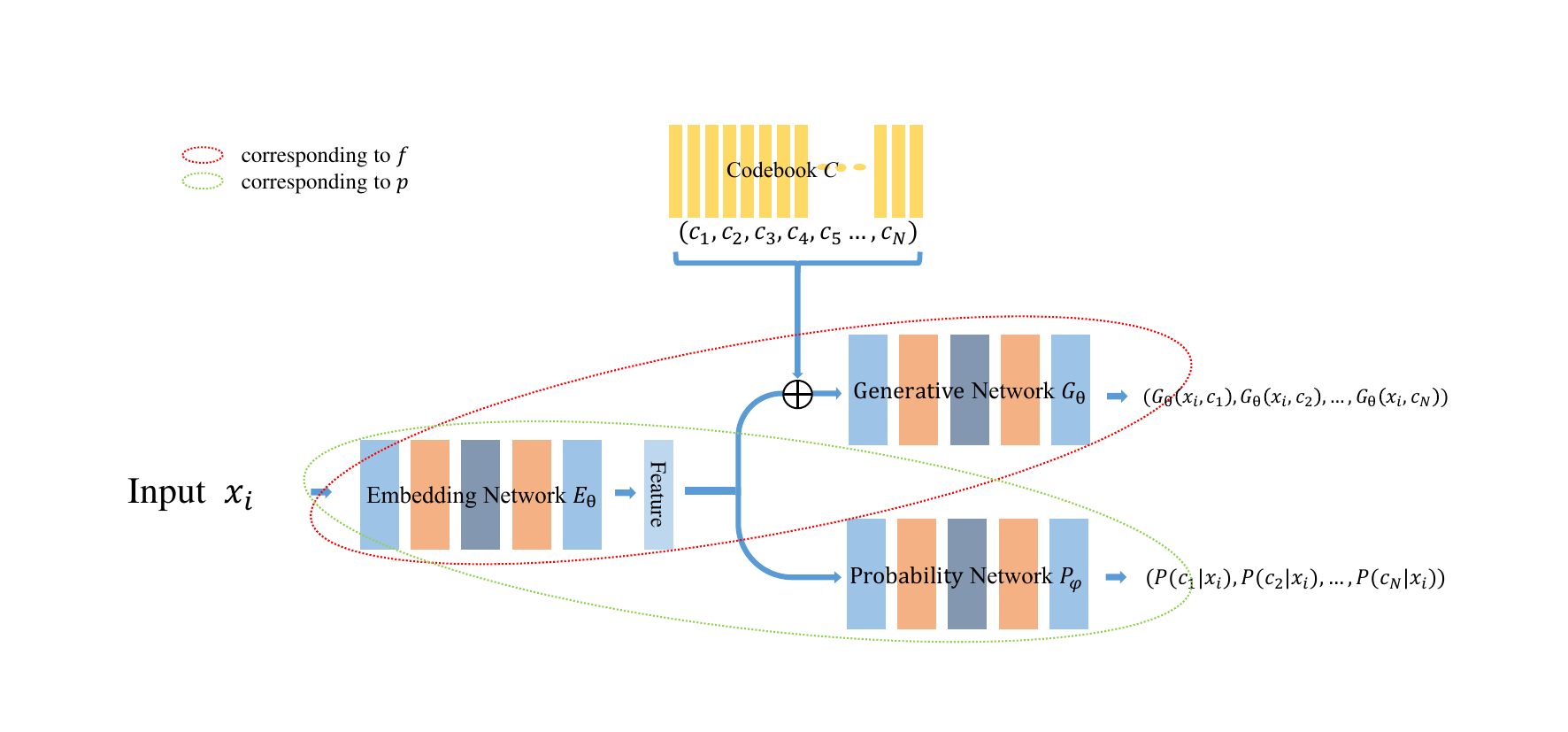}
	\caption{The architecture of our framework for modeling DMM.}
        \label{dmm_framework}
\end{figure}

Our objective is to develop a deep neural network framework to solve the optimization problem \ref{final_loss}. To achieve this, we parameterize the mappings $f$, $p$, and $z$ using deep neural networks $G_{\theta}, P_{\varphi}$ and $E_{\phi}$ respectively. The network structures are illustrated in the Fig. \ref{dmm_structure}. The framework for formulating the DMM is shown in Fig. \ref{dmm_framework}.

The deep neural network $G_{\theta}$ represents the bi-variate mapping $f$. In other words, we have $f(x,c_j)= G_{\theta}(x,c_j)$. It is depicted within the region bounded by the red dotted boundary in Fig. \ref{dmm_structure}. The input $x \in X$ is passed through an embedding network, which generates a latent representation $l$. This latent representation captures the meaningful features of $x$. Subsequently, the latent representation, along with the codebook $C = \{c_1, c_2, ..., c_N\}$, is fed into another deep generative network that produces plausible outputs ${f(x, c_1), f(x, c_2), ..., f(x, c_N)}$. Here, we assume that the discrete codebook $C$ captures shared label information across different $x \in X$. It is worth noting that different parameters $\theta$ of the deep neural network result in different one-to-$N$ mappings. Within this framework, we can effectively search for the optimal $f$ by optimizing $\theta$.

Similarly, the deep neural network $P_{\varphi}$ represents the bi-variate mapping $p$. In other words, $p(c_j|x) = P_{\varphi}(c_j,x)$. It is shown within the region bounded by the green dotted boundary in Fig. \ref{dmm_structure}. The input $x \in X$ is first passed through an embedding network, which produces a latent vector $h$. By performing a matrix multiplication $Ah$ and applying a softmax operation, we obtain a probability vector $p = (p(c_1|x), p(c_2|x), ..., p(c_N|x))$. The output of $P_\theta$ estimates the probability $p(c_j| x)$ for each plausible output $f(x, c_j)$. The choice of the fixed matrix $A$ is crucial for accurately predicting uncertainty estimation, and we will discuss this in detail later. In addition, we fuse the embedding networks for $G_{\theta}$ and $P_{\theta}$ as one named $E_{\theta}$. This reduces the parameters of the model and improves the efficiency of training. For simplification, we abbreviate $G_{\theta}(E_{\theta}(x),c_j)$ and $P_{\varphi}(c_j,E_{\theta}(x))$ to $G_{\theta}(x,c_j)$ and $P_{\varphi}(c_j,x)$ respectively.

To solve the optimization problem \ref{final_loss}, we introduce the cluster mapping $z$. The cluster mapping takes $x\in X$ and an associated plausible output $y\in Y$ from the dataset as the input and output a vector, which is of the same dimension of the code in the codebook. To parameterize $z$, we utilize another deep neural network $E_{\phi}$. In other words, $z(x,y) = E_{\phi}(x,y)$. As shown within the region bounded by the blue dotted boundary in Fig. \ref{dmm_structure}, $x\in X$ and $y\in Y$ are fed into an embedding network to output a vector $E_{\phi}(x,y)$. With $E_{\phi}(x,y)$, we can find the code in $C$ closest to $E_{\phi}(x,y)$ that solves \ref{closestcode}.

Under this setting, all mappings $f$, $p$, and $z$ to be optimized are parameterized using deep neural networks. Therefore, they can be optimized by training the deep neural networks to obtain the optimal parameters that minimize a loss function defined by the energy functional in our optimization problem.

\subsubsection{Optimization of discrete codebook $C$}
A crucial component of our proposed framework is the use of a discrete codebook $C = \{c_1, c_2, ..., c_N 
\} \subset \mathbb{R}^m$ to represent a one-to-$N$ mapping. For each $x \in X$, we can obtain $N$ plausible outputs: $f(x, c_1)$, $f(x, c_2)$, ..., $f(x, c_N)$. Additionally, we can estimate the corresponding probabilities associated with each output: $p(c_1, x)$, $p(c_2, x)$, ..., $p(c_N, x)$.

It is important to highlight that for our framework to effectively represent a DMM, the following condition must hold for any $j \neq k$: if both $p(c_j, x)$ and $p(c_k, x)$ are non-zero, then $f(x, c_j) \neq f(x, c_k)$. In other words, each code in the codebook should be associated with a distinct output for every $x \in X$. The choice of the codebook is therefore crucial in enforcing this requirement.

In our framework, for every $x \in X$, a code $c_j \in C$ corresponds to a plausible output $G_{\theta}(x, c_j)$. It is important to note that $c_j \in \mathbb{R}^m$, where usually the dimension $m$ of $c_j$ is significantly smaller than the dimension of the output space $Y$. Consequently, the generator produces plausible outputs $G_{\theta}(x, c_j) \in Y$ that have a much higher dimension. The separability of the codes $c_j$ encourages the separability of the corresponding outputs $f(x, c_j)$. Conversely, if two codes $c_j$ and $c_k$ are close to each other in the codebook $C$, the plausible outputs $f(x, c_j)$ and $f(x, c_k)$ will also be close to each other. This can hinder the capability of our framework to effectively represent a DMM.

In practice, although the discrete codebook inherently lends itself to modeling multi-modal label data $\{(x_i, y_i^k)_{k=1}^{\mathcal{N}_i}\}_{i=1}^{T}$, the occurrence of similar codes within the codebook during continuous updates can lead to highly similar results. To prevent this repetition and ensure diversity in the generated outcomes, it becomes crucial to maximize the separation between each code in the codebook $C$. 

To maximize the separation between each code in the codebook, our strategy is to reduce the mutual correlation among vectors in the codebook. For this purpose, we introduce the following {\it covariance loss}.
\begin{definition}
Let $\mathcal{X} = \{x_1,x_2,...,x_N\}\subset \mathbb{R}^m$ be a finite subset of $\mathbb{R}^m$ with $|x_1| = ...= |x_N|=1$. The {\it covariance loss} with a threshold $\tau$ is defined as
\begin{equation}
    L_{cov}^{\tau}(\mathcal{X}) = \frac{||T^{\tau}(\mathcal{X}^T \mathcal{X} - I_N)||_F^2}{\mathbb{I}\{T^{\tau}(\mathcal{X}^T \mathcal{X} - I_N)\}}
\end{equation}
\noindent where $||\cdot||_F$ is the Frobenius norm,  $T^{\tau}:\mathbb{R}^{N\times N}\to \mathbb{R}^{N\times N}$ given by:
\begin{equation}
T^{\tau}(M)=
\begin{cases}
M_{ij}, \text{ if } |M_{ij}|> \tau;\\
0, \text{ otherwise},
\end{cases}
\end{equation}
\noindent where $M\in \mathbb{R}^{N\times N}$ and $M_{ij}$ is the $i$-th row $j$-th column entry of $M$. Thus,  Also, $\mathbb{I}:\mathbb{R}^{N\times N}\to \mathbb{N}$, where $\mathbb{I}(M)$ is the number of non-zero entries of $M$.
\end{definition}

To understand the meaning of the covariance loss, observe that all diagonal entries of $\mathcal{X}^T \mathcal{X}\in \mathbb{R}^{N\times N}$ are equal to 1. Thus, $\mathcal{X}^T \mathcal{X}- I_N$ has 0 on its diagonal. Each non-diagonal entry of $\mathcal{X}^T \mathcal{X} - I_N$ represents the inner product between two distinct data points in $\mathcal{X}$. By applying a threshold $\tau$ to $T^{\tau}(\mathcal{X}^T \mathcal{X} - I_N)$, we retain only the entries that exceed $\tau$. These entries signify pairs of data points that are close to each other, as their inner product surpasses the threshold. The covariance loss measures the mean squared sum of all non-zero entries in $T^{\tau}(\mathcal{X}^T \mathcal{X} - I_N)$. Minimizing the covariance loss $L_{cov}^{\tau}$ aims to encourage a greater separation between the data points in $\mathcal{X}$. By this loss function, we are only concerned with those vectors that are not almost orthogonal, i.e., inner products greater than the threshold. It can help us focus on preventing the occurrence of similar codes within the selected codes. In particular, when $L_{cov}^{\tau} = 0$, it implies that the inner product between any pair of data points in $\mathcal{X}$ is below the threshold $\tau$. This indicates that the separability of each pair of data points in $\mathcal{X}$ satisfies a set tolerance, promoting greater distinctiveness among the data points.

Using $L_{cov}^{\tau}$, we can encourage the separation of the codes in the codebook $C = \{c_1, c_2, ..., c_N\}$. To achieve this, we first normalize the codes in the codebook to $\widetilde{C} = \{\frac{c_1}{|c_1|}, \frac{c_2}{|c_2|}, ..., \frac{c_N}{|c_N|}\}$. By minimizing the covariance loss $L_{cov}^{\tau}(\widetilde{C})$, we can effectively separate the normalized codes in the codebook $C$. This, in turn, promotes the separability of the corresponding plausible outputs $f(x, c_j)$ in the output space $Y$, ensuring a diverse set of generated results for each input $x$.

Furthermore, choosing suitable parameters for the size of the codebook $N$ and the dimensionality of the codes $m$ is also important. For a set of unit vectors $\{v_1, ..., v_N\}$ in $\mathbb{R}^m$, if $N \leq m$, it is possible for their inner products to be all equal to zero, ensuring perfect orthogonality. However, in practical applications, the hyperparameters $N$ and $m$ of the codebook may require a broader range of choices to adapt to different datasets. It is not always the case that $N \leq m$. In situations where $N > m$, we require the variables in the codebook to be "almost orthogonal." The Kabatjanskii-Levenstein bound for almost orthogonal vectors \cite{tao2013almostorthogonal} helps us to decide on appropriate choices for the values of $N$, $m$, and the threshold $t$ in such cases. This theorem provides theoretical guidance on the trade-offs between the codebook size $N$, the code dimensionality $m$, and the degree of orthogonality required, allowing us to configure these parameters effectively for different datasets.

\begin{theorem}
    Let $v_1, v_2, ..., v_m$ be unit vectors in $\mathbb{R}^n$ such that $\|<v_i,v_j>\| \leq An^{-1/2}$ for all distinct $i,j$, $\frac{1}{2} \leq A \leq \frac{\sqrt{n}}{2}$, then we have $m\leq(\frac{Cn}{A^2})^{CA^2}$ for some absolute constant $C$.
\end{theorem}

For the special case when the hyperparameter $A=1/2$, we have the following theorem \cite{tao2013almostorthogonal}:

\begin{theorem}
Let $v_1,v_2,...,v_m$ be unit vectors in $\mathbb{R}^n$ such that $|<v_i,v_j>|\leq \frac{1}{2n^{1/2}}$ for all distinct $i,j$. Then, $m<2n$.
\end{theorem}

Therefore, in our case, we set the hyperparameter $A=1/2$. Then, the threshold $\tau$ is chosen as $1/2m^{-1/2}$. As a condition on the hyperparameter choices, we must have $N < 2m$, where $N$ is the size of the codebook and $m$ is the dimensionality of the codes. In our work, we carefully select the values of $N$ and $m$ such that the separability tolerance can be achieved using this threshold. With these parameters, we will incorporate the thresholded covariance loss $L_{cov}^{\tau}$ in the overall loss function for training the deep neural network to solve the optimization problem in \ref{final_loss}. Our experimental results demonstrate the powerful impact of the covariance loss in improving the separability and diversity of the generated outputs.

In practice, the number of codes $N$ is often larger than the actual number of plausible outputs. This means that $\mathcal{N}_{x} \ll N$ for all $x \in X$, resulting in some codes being unused and left idle. Initially, the codes are configured such that the inner products between distinct pairs are less than the threshold. However, when updating the codes that have been used, they tend to accumulate in similar positions, reducing their separability. Minimizing the covariance loss plays a crucial role in addressing this issue. When minimizing the covariance loss, only the active codes are modified, while the inner products involving non-active codes remain at 0 even after applying the threshold function $T^{\tau}$. This ensures that the focus is on adjusting the positions of the codes that are actually contributing to the plausible outputs, enhancing their separability and promoting diversity in the output space.

Another important challenge to address is that when solving the optimization problem \ref{final_loss}, the following condition \ref{closestcode} has to be considered:
\[
c_{s_{i}^{k}} = \mathop{\arg\min}\limits_{c \in C} ||(E_{\phi}(x_i, y_{i}^{k})- \ c||^2
\]

During the training process, a data-label pair $(x_i, y_i^{k})$ is randomly sampled from the unorganized dataset ${(x_i,y_i^{k})}_{k=1}^{\mathcal{N}_i}$, and fed into the embedding network $E_{\phi}$. Subsequently, the nearest code $c$ is selected from the codebook for the corresponding feature $E_{\phi}(x_i, y_i^{k})$, as follows:

\begin{equation}
    c(x_i,y_i^{k}) = c_{s_{i}^{k}}, \ \text{where } \ s_{i}^{k} = \mathop{\arg\min}\limits_{s=1,..., N}\|c_{s} - E_{\phi}(x_i,y_i^{k})\|^2.
    \label{nearest}
\end{equation}

Note that the selected code $c(x_i, y_i^k)$ replaces the original feature $E_\phi(x_i, y_i^k)$ as the input for the generative network $G_\theta$. However, this approach presents a potential challenge: there is no direct gradient of $E_\phi(x_i, y_i^k)$ from the data fidelity term in the backward propagation process.
To address this problem, we use a simple gradient approximation method, following the approach of \cite{van2017neural}. The key idea is to copy the gradient of the selected code $c$ and assign it to the feature $E_\phi(x_i, y_i^k)$ so that the parameters $\phi$ of the embedding network can be updated using the gradient information from the loss function. Specifically, in the forward pass, we directly input the feature representation $c$ to the generator $G_\theta$. In the backward computation, we directly assign the gradient $\nabla_c L$ to the embedding network $E_\phi$. To ensure that this gradient approximation is meaningful, we need to make the output $E_\phi(x_i, y_i^k)$ as close as possible to the selected code $c$. To achieve this, we add an extra regularization loss for $E_{\phi}$ to the overall loss function, given by:
\[
E_{zreg}(\phi) = \beta \sum_{i=1}^T \sum_{k=1}^{\mathcal{N}_i} |E_\phi(x_i, y_i^k) - \text{sg}[c(x_i,y_i^k)]|^2,
\]
\noindent where $\beta > 0$ and $\text{sg}$ is the stop-gradient operation (identity in the forward pass, zero derivative in the backward pass, and easily implemented in neural network algorithms). This regularization loss encourages the embedding network to output features that are close to the selected codes, without backpropagating gradients to the codebook $C$ itself.

The rationale behind this approach is that a learnable codebook, while exhibiting a slower learning capability, is required to capture more information from the data. Therefore, we propose to update the codebook $C$ using a dictionary learning algorithm, Vector Quantization (VQ) \cite{van2017neural}, which computes the exponential moving average of the corresponding embedding network's outputs. Our ablation analysis \ref{variation} has shown that a learnable codebook outperforms a fixed codebook.

\subsubsection{Probability prediction}

Another crucial component in our framework is the estimation of the probability associated with each plausible output. In practice, the ground truth probability of each plausible output associated with an input is not known. Our goal is to estimate these probabilities from the sample training dataset $\{\{(x_i,y_i^{k})\}_{k=1}^{\mathcal{N}_i}\}_{i=1}^T$. For each input $x_i$, recall that the collection of sampled plausible outputs $\{y_i^k\}_{k=1}^{\mathcal{N}_i}$ can contain repeated values. For example, when labeling lesions of a medical image, different medical experts might provide the same label, resulting in repeated labels in the dataset. The more repetitions of a particular label, the higher the estimated probability associated with that plausible output. The intuition behind this approach is that the frequency of a plausible output in the sample dataset can serve as a proxy for its true probability. The more often a plausible output appears, the more likely it is to be the correct label for the given input. By leveraging this idea, we can estimate the probability distribution over the plausible outputs for each input, even though the ground truth probabilities are not known.

In our framework, the probability $p: C \times X \to [0, 1]$ is parameterized by a deep neural network $P_\varphi$. Given an input $x_i$ from the training dataset, it is fed into the network to obtain a feature vector $h \in \mathbb{R}^m$.
Next, we compute a vector $\tilde{p}_i \in \mathbb{R}^N$ as $\tilde{p}_i = Ah$, where $A \in \mathbb{R}^{N \times m}$ is a suitable matrix. We then pass $\tilde{p}_i$ through a softmax operation to obtain the probability vector $p_i$, where:
\begin{equation}
    p_{ij} = \frac{e^{\tilde{p}_{ij}}}{\sum_{j=1}^m e^{\tilde{p}_{ij}}}.
\end{equation}

\noindent Here, $p_{ij}$ represents the probability of the plausible output $f(x_i, c_j)$. If $p_{ij} = 0$, the output $f(x_i, c_j)$ is considered meaningless and can be ignored.

The choice of the matrix $A$ is crucial in this setup. The value of $p_{ij}$ is related to the inner product between the $j$-th row of $A$ and the vector $\tilde{p}_i$. Specifically, $p_{ij}$ will be larger if $\tilde{p}_i$ is closer to the $j$-th row of $A$. One possible approach is to learn the matrix $A$ during the training process, allowing the network to discover the optimal projection that captures the relationship between the input features and the plausible output probabilities.

However, when training on an imbalanced dataset, the learnable vectors of the minority classes may collapse, a phenomenon known as ``minority collapse". To alleviate this issue, we define $A$ using the simplex equiangular tight frame (ETF) \cite{papyan2020prevalence}. This results in a fixed ETF classifier, as proposed in \cite{yang2022we}.

The simplex equiangular tight frame (ETF) is formally defined as follows:

\begin{definition}{(Simplex Equiangular Tight Frame)}\label{ETF}
A collector of vectors $m_i \in \mathbb{R}^m, i=1,2,..., N, m\geq N-1$, is said to be a {\it simplex equiangular tight frame} if:
\begin{equation*}
    M = \sqrt{\frac{N}{N-1}}U(I_N - \frac{1}{N}1_{N}1_{N}^{T}),
\end{equation*}
where $M=[m_1,...,m_N]\in\mathbb{R}^{m\times N}$,$U\in\mathbb{R}^{m\times N}$ allows a rotation and satisfies $U^{T}U=I_N$, $I_N$ is the identity matrix, and $1_N$ is an all-ones vector.
\end{definition}

All vectors in a simplex ETF have an equal $l_2$ norm and the same pair-wise angle, i.e.:
$m_i^T m_j = \frac{N}{N-1}\delta_{i,j} - \frac{1}{N-1}, \quad \forall i,j \in {1,2,...,N}$
where $\delta_{i,j}$ equals 1 when $i=j$ and 0 otherwise. The pair-wise angle $-\frac{1}{N-1}$ is the maximal equiangular separation of $N$ vectors in $\mathbb{R}^m$ \cite{papyan2020prevalence}.

We can then define $A$ whose $j$-th row is given by $m_j$. To find the optimal $p$, we optimize the parameters of $P_{\varphi}$ given a training dataset $\{\{(x_i,y_i^{k})\}_{k=1}^{\mathcal{N}_i}\}_{i=1}^T$ to minimize $E_3$ in our optimization problem \ref{final_loss}. In particular, the first term in $E_3$ involves $P_{\varphi}$:
\begin{equation}
\begin{aligned}
            \frac{1}{\mathcal{T}}\sum^{\mathcal{T}}_{i=1} \frac{1}{\mathcal{N}_i}\sum^{\mathcal{N}_i}_{k=1} \log \left(\frac{1}{p(c_{s_{i}^{k}} | x_i)^{\alpha}}d(f(x_i, c_{s_{i}^{k}}), \ y_{i}^{k}) \right) = &\frac{1}{\mathcal{T}}\sum^{\mathcal{T}}_{i=1} \frac{1}{\mathcal{N}_i}\sum^{\mathcal{N}_i}_{k=1} \log (d(f(x_i, c_{s_{i}^{k}}), \ y_{i}^{k})) \\
            &- \alpha\log(P_\varphi(c_{s_{i}^{k}},x_i))
\end{aligned}
\end{equation}

Therefore, $P_\varphi$ can be optimized by minimizing the following cross-entropy loss:
\[
L_{CE}(h,M) =  \frac{1}{\mathcal{T}}\sum^{\mathcal{T}}_{i=1} \frac{1}{\mathcal{N}_i}\sum^{\mathcal{N}_i}_{j=1} \left( -log (\frac{exp(h^T m_{s_{i}^{j}})}{\sum_{k=1}^N \exp(h^T m_k)})\right),
\]
where $M=[m_1,...,m_N]\in\mathbb{R}^{m\times N}$ is the fixed ETF classifier generated by Definition \ref{ETF}.

\subsubsection{Loss Function} 
To solve the optimization problem \ref{final_loss}, our framework is reduced to finding optimal parameters $\theta, \varphi$, and $\phi$, which minimizes $E_3$. More specifically, $E_3$ can now be written as follows:

\begin{equation}
\begin{aligned}
 E_3 ({\theta,\varphi,\phi, C}) = \frac{1}{\mathcal{T}}\sum^{\mathcal{T}}_{i=1} \frac{1}{\mathcal{N}_i} &\sum^{\mathcal{N}_i}_{k=1} \log \left(\frac{1}{P_{\varphi}(c_{s_{i}^{k}} , x_i)^{\alpha}}d(G_\theta(x_i, c_{s_{i}^{k}}), \ y_{i}^{k}) \right) \\
 &+ \beta d''(E_\phi(x_i, y_{i}^{k}), c_{s_{i}^{k}}) + \gamma d'''(C)   
\end{aligned}
\end{equation}

As discussed in the previous subsections, the regularization term $d''$ for $E_\phi$ is chosen as $d''= E_{zreg}$. Also, the regularization for $C$ is chosen as $d'''= L_{cov}^{\tau}$. Hence, 
\begin{equation}
E_3 ({\theta,\varphi,\phi, C}) =  L_{recon}(\theta) + \alpha L_{CE}(h, M) + \beta E_{zreg}(\phi) + \gamma L_{cov}^{\tau}(C)
\end{equation}
\noindent where 
\begin{equation}
    L_{recon}(\theta) = \frac{1}{\mathcal{T}}\sum^{\mathcal{T}}_{i=1} \frac{1}{\mathcal{N}_i}\sum^{\mathcal{N}_i}_{j=1} \log d(G_{\theta}(x_i, c(x_i,y_i^k)), \ y_{i}^{k}).
\end{equation}

\noindent The choice of distance function $d$ depends on datasets and tasks. The overall loss function to train the deep neural network to solve the optimization problem \ref{final_loss} can now be summarized as follows.

\begin{equation}
\label{train_loss}
\begin{split}
        L(\theta, \varphi,\phi, C) = & \frac{1}{\mathcal{T}}\sum^{\mathcal{T}}_{i=1} \frac{1}{\mathcal{N}_i}\sum^{\mathcal{N}_i}_{j=1} \log d(G_{\theta}(x_i, c(x_i,y_i^k)), \ y_{i}^{k}) -\alpha \log(P_{\varphi}(c(x_i,y_i^k),x_i)) \\
        & +\beta \|E_{\phi}(x_i,y_i^{k}) - sg[c(x_i,y_i^k)]\|^2 + \gamma  \frac{||T^{\tau}(\tilde{C}^T \tilde{C} - I_N)||_F^2}{\mathbb{I}\{T^{\tau}(\tilde{C}^T\tilde{C} - I_N)\}}
    \end{split}
\end{equation}

The parameters of the deep neural network can then be optimized by stochastic gradient descent through backward propagation.

\subsubsection{Numerical algorithm}

We describe the numerical algorithms in detail. Several techniques, such as simple gradient approximation, stop-gradient operation, and exponential moving average, are used in our numerical algorithms.

Firstly, the simple gradient approximation is used during the update of $\phi$. Since the encoder $E_{\phi}$ receives no gradient from the reconstruction loss, we assign it the gradient of the code $c(x_i,y_i^{k})$, abbreviated as $c$ here. Thus, with the input data pair $(x_i,y_i^{k})$, we have:

\begin{equation}
\label{loss_gradient}
\begin{aligned}
\partial_{\phi}L(\theta, \varphi, \phi, C) &= \frac{\partial L_{recon}(G_{\theta}(x_i,c),y_i^{k})}{\partial c} \cdot \frac{\partial E_{\phi}(x_i,y_i^{k})}{ \partial \phi} + \frac{\beta \cdot \partial \|E_{\phi}(x_i,y_i^{k}) - sg[c]\|^2}{\partial \phi}\\
\partial_{\theta}L(\theta, \varphi, \phi, C) &= \partial_{\theta} L_{recon}(G_{\theta}(x_i,c),y_i^{k}) \\
\partial_{\varphi}L(\theta, \varphi, \phi, C) &=  \alpha \cdot \partial_{\varphi}L_{CE}(h, M)\\
\partial_{C}L(\theta, \varphi, \phi, C) &= \gamma \cdot \partial_{C} L_{cov}(C).\\
\end{aligned}
\end{equation}

Here, the first equation approximates the gradient of the loss function with respect to the encoder parameters $\phi$. The second equation computes the gradient with respect to the generator parameters $\theta$, which includes both the reconstruction loss and the cross-entropy loss. The third equation computes the gradient with respect to the code matrix $C$.

Note that the selected codes receive no gradient from the reconstruction loss and regularization loss, but instead from the covariance loss. In addition, the codes are also updated by the exponential moving average \cite{van2017neural} as follows:
\begin{equation}
\label{exp_moving}
    \begin{aligned}
        n^{t} &= n^{t-1} * \kappa + count^{t} * ( 1- \kappa) \\
        m^{t} &= m^{t-1} * \kappa + \sum^{count^{t}}_{j} e_j^{t} * ( 1- \kappa) \\
        c^{t} &= \frac{m^{t}}{n^{t}},
    \end{aligned}
\end{equation}
where $\{e_1, e_2, ..., e_{count}\}$ is a set of the embedding features that are closest to code $c$. $0< \kappa <1$ is the decay coefficient.

During the testing process, the probabilities predicted by $P_{\theta}$ are not exactly equal but very close to 0. A small threshold $\epsilon = 1e-5$ is set to eliminate results with extremely low probability.

The details of the numerical algorithms for the training and testing processes are described in Algorithms \ref{algorithm_1} and  \ref{algorithm_2}, respectively.

\begin{algorithm}
    \caption{Training process of DMM Framework.}
    \label{algorithm_1}
    \KwIn{weights $\alpha$, $\beta$, $\gamma$ in loss function \ref{train_loss}, parameter $N, m$ in codebook, learning rate $lr$.}
    \KwOut{network parameters $\theta$, $\varphi$, $\phi$ and codebook $C$.}
    \BlankLine
    Initialize embedding network $E_{\theta}$, $E_{\phi}$, generator $D_{\theta}$, classifier $P_{\varphi}$ and codebook $C$.\\
    \For{$t=0,1,2,..$}{
    Sample batch of data pair $(x_i,y_i^{k})$ from dataset $\{\{(x_i,y_i^{k})\}_{k=1}^{\mathcal{N}_i}\}_{i=1}^T$ randomly.\\
    Compute nearest $c_{s_i^k}$ by \ref{nearest}.\\
    Compute $\partial_{\theta}L(\theta^{t}, \varphi^{t}, \phi^{t}, C^{t})$, $\partial_{\varphi}L(\theta^{t}, \varphi^{t}, \phi^{t}, C^{t})$, $\partial_{\phi}L(\theta^{t}, \varphi^{t}, \phi^{t}, C^{t})$, and $\partial_{C}L(\theta^{t}, \varphi^{t}, \phi^{t}, C^{t})$ by \ref{loss_gradient}.\\
    Update $c_{s_i^k}$ by the exponential moving average \ref{exp_moving}.\\
    Update $\theta$, $\varphi$, $\phi$, and $C$ by\\
    \qquad $\theta^{t+1} = \theta^{t} + lr \cdot \partial_{\theta}L(\theta^{t}, \varphi^{t}, \phi^{t}, C^{t})$\\
    \qquad $\varphi^{t+1} = \varphi^{t} + lr \cdot \partial_{\varphi}L(\theta^{t}, \varphi^{t}, \phi^{t}, C^{t})$\\
    \qquad $\phi^{t+1} = \phi^{t} + lr \cdot \partial_{\phi}L(\theta^{t}, \varphi^{t}, \phi^{t}, C^{t})$\\
    \qquad $C^{t+1} = C^{t} + lr \cdot \partial_{C}L(\theta^{t}, \varphi^{t}, \phi^{t}, C^{t})$.\\   
    }
    \Return{$\theta$, $\varphi$, $\phi$ and $C$}.
\end{algorithm}

\begin{algorithm}
    \caption{Testing process of DMM Framework.}
    \label{algorithm_2}
    \KwIn{$x \in X$.}
    \KwOut{multi-labels $y_1, y_2, ..., y_{\mathcal{N}_{x}}$ and their probabilities $P(y_1|x), P(y_2|x), ...,P(y_{\mathcal{N}_{x}}|x)$.}
    
    Sample a test data $x$.\\

    Compute $\{P(\hat{c}_1|x), P(\hat{c}_2|x), ...,P(\hat{c}_{N}|x)\}$ by $P_{\theta}(x)$.\\
    $\mathcal{N}_{x} = N - count\{P(\hat{y_j}|x) < \epsilon\}$.\\
    
    Obtain $\{\hat{c}_1, \hat{c}_2, ...,\hat{c}_{\mathcal{N}_{x}}\}$ by reordering $\{c_1, c_2, ..., c_N\}$ in descending probabilities while removing items whose $P(\hat{c}_j|x) < \epsilon$.\\
    
    \For{$j=1, 2, ..., \mathcal{N}_{x}$}{
    $y_j = D_{\theta}(x, \hat{c}_j).$}

    \Return{$\{y_1, y_2, ..., y_{\mathcal{N}_{x}}\}$, $P(y_1|x), P(y_2|x), ...,P(y_{\mathcal{N}_{x}}|x)$}.
\end{algorithm}

\section{Experiments}\label{sec_exp}
To evaluate the effectiveness of our proposed framework, we conducted experiments on both synthetic examples and real-world imaging problems. We also performed ablation studies to analyze the key components of the framework. In this section, the experimental results will be reported.

\subsection{Experimental setup}
We utilize convolutional neural networks (CNNs) for both the embedding networks and the generative module, resembling the U-Net architecture as in \cite{ronneberger2015u}. To be more precise, both our embedding networks $E_{\phi}$ and $E_{\theta}$ consist of a sequence of downsampling residual blocks, while the generative module $G_{\theta}$ is comprised of a sequence of upsampling residual blocks. Additionally, the generative network incorporates feature information from the embedding network $E_{\theta}$ at each resolution level. There are four downsampling or upsampling blocks in the sequence of each module. The downsampling and upsampling operations use bilinear interpolation. Each residual block comprises three convolution layers, utilizing $3 \times 3$ kernels and ReLU activation. The two embedding networks have the same architecture with output channel dimension $[32, 64, 128, 256]$. The 1×1 convolution and global average pool follow the network $E_{\phi}$ of the data-label pair to obtain the feature of the same dimension as the code in the codebook. However, a fixed ETF classifier as a probability network follows the embedding network $E_{\theta}$ to output the probability prediction of all codes in the codebook. The categories of the classifier are the same size as the codebook. We incorporate the selected code and the features of the generative network to the model's last layers with $1 \times 1$ convolutions and finally activated by softmax. Note that the code can incorporate any skip connection to the generative module $G_{\theta}$ for different tasks. Since the code selected from the codebook is a vector, it is impossible to directly incorporate the code into a generation network with spatial dimensions. We repeatedly extend each value in the code to the spatial size of the corresponding features, and then concatenate them. We initialize the codebook $C$ as a $(256, 256)$ matrix of i.i.d random rotation matrices. Each column of the codebook represents an individual code with a norm of 1. During the training, we utilized the binary cross-entropy loss as label reconstruction loss and set the batch size to $32$. Additionally, we employed a learning rate schedule with values of $[1e^{-4}, 5e^{-5}, 1e^{-5}, 5e^{-6}]$ at epochs $[0, 300, 900, 1200]$. The $l^2$ penalization weight was fixed at $\beta = 0.25$, another weight of covariance constraint is $\gamma = 0.01$, and we utilized the Adam optimizer \cite{kingma2014adam} with its default settings for all of our experiments. Specifically, we train the first $20$ epochs without the cross-entropy loss to avoid the impact of violent fluctuations in early code selection on the learning of probability network parameters. For the Probabilistic U-Net, we followed the parameter $num_{filter} = [32, 64, 128, 192]$ in its released version and the suggested hyperparameters for segmentation tasks in \cite{kohl2018probabilistic}.

\subsection{Synthetic examples: Shape reconstruction} 

To rigorously evaluate the capabilities of our proposed framework, we first conducted experiments on a synthetic dataset for the task of shape reconstruction. We generated a dataset of shape data-label pairs, where the data consisted of a randomly generated triangle image of size (160, 160), and the labels represented four distinct shapes derived from the properties of the input triangle.

Specifically, the dataset generation process was as follows. First, we randomly created a triangle image to serve as the input data. We then synthesized four corresponding label images, each representing a shape related to, but distinct from, the original triangle:
\begin{enumerate}
    \item The first label was a smaller triangle created by cropping a random portion of the input triangle.
    \item The second label was the original input triangle.
    \item The third label was a pentagon shape formed by cutting a random triangle from the upper-right corner of the parallelogram constructed from the input triangle.
    \item The fourth label was a complete parallelogram shape generated by extending the parallelogram formed by the input triangle.
\end{enumerate}

So the label shapes were not simply transformations of the input triangle, but rather new shapes that were algorithmically derived from the properties and geometry of the original triangle. This allowed us to evaluate how well our model could capture the underlying relationships between the input triangle and these related shape outputs.

Using this procedure, we constructed a synthetic dataset consisting of 2,000 data-label pairs for the training set and 200 pairs for the test set. 
Our goal is to find an optimal DMM that fits the training dataset. Given an input triangle, the optimal DMM should predict all four shapes, along with an estimation of the probability associated with each output shape.

In the training process, we repeatedly randomly sample a triangle $x_i$ from the training set and one of its four labels as $y_{i}^{k}$. Since the four labels are different, random sampling ensures that the distribution of the label space consists of four modes with the same probability equal to 0.25.

\begin{figure*}[t]
	\centering
	\includegraphics[width=2.8in]{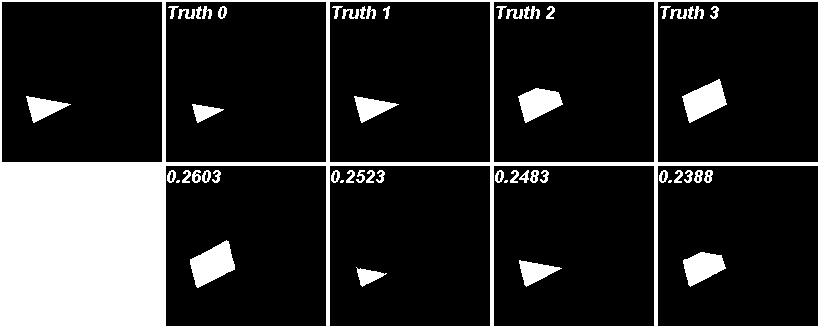}
    \includegraphics[width=2.8in]{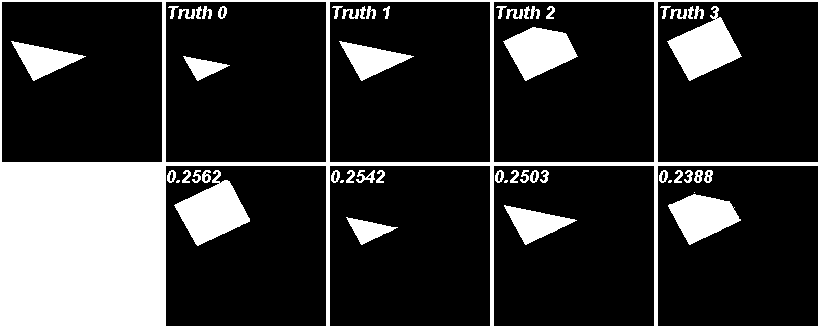}
	\caption{Results visualization for shape reconstruction. The first row shows the input samples and their labels, and the next row shows the predictions from our method. The probability for each prediction is annotated in the upper-left corner.}
	\label{shape} 
\end{figure*}

In the testing phase, we sample a triangle from the testing data and record its most likely outputs ${\hat{y}_i^j | P(\hat{y}_i^j|x_i) > 1e^{-5}, j=1,..., N }$. Fig. \ref{shape} shows the results of two input triangles. In each case, the leftmost image in the first row shows the input sample. The other four images show the 4 labeled shapes associated with the input triangle. Note that the input and output pairs in the testing have not been used in the training process.

The second row of Fig. \ref{shape} shows the predictions by the optimal DMM. Observe that the generated plausible outputs closely resemble the 4 ground truth shapes. This demonstrates the efficacy of our proposed framework to make multi-modal predictions. The value in the top left corner of each output shows the probability associated with each plausible output generated by the DMM. All values are very close to the ground truth probability of 0.25. This demonstrates that our framework is successful in obtaining accurate uncertainty estimates corresponding to each output.

\begin{figure*}[t]
	\centering
	\includegraphics[width=2.8in]{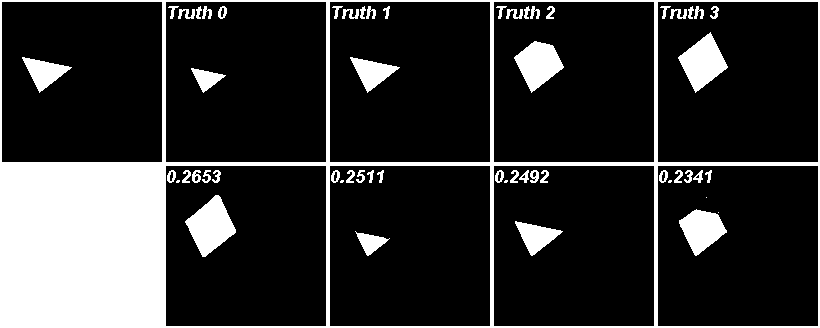}
    \includegraphics[width=2.8in]{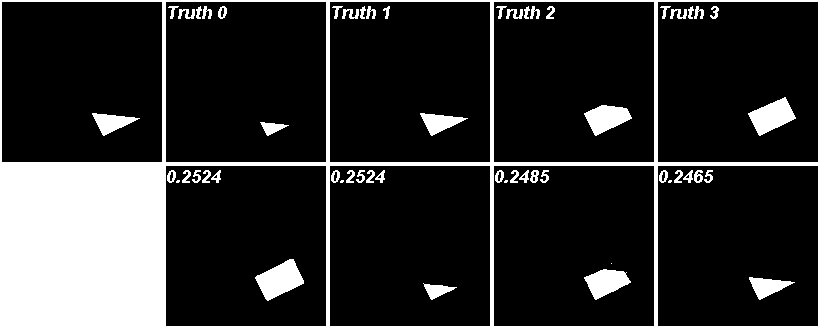}\\
	\includegraphics[width=2.8in]{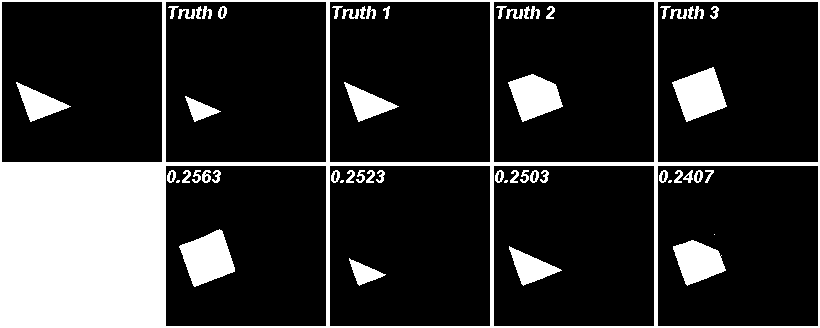}
    \includegraphics[width=2.8in]{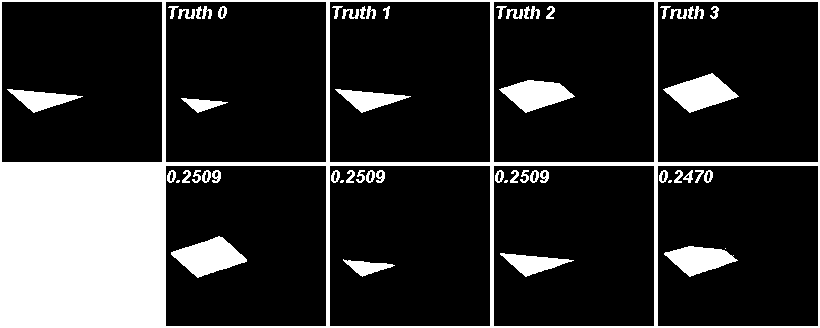}\\  
    \includegraphics[width=2.8in]{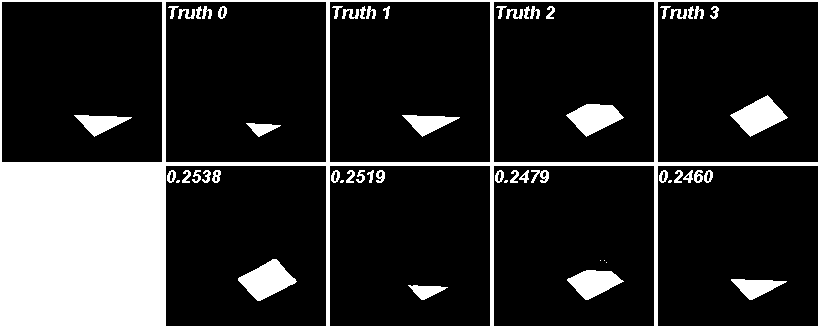}
    \includegraphics[width=2.8in]{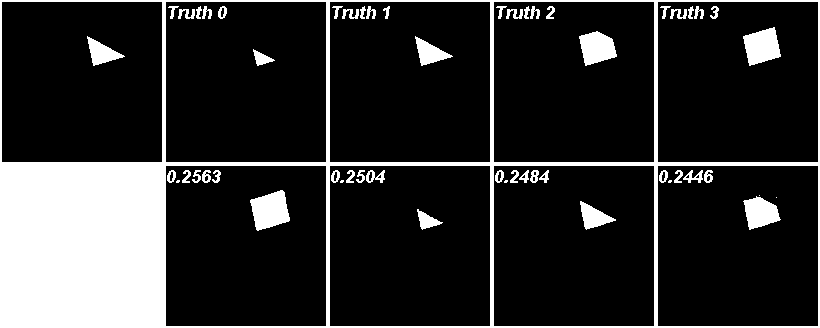}\\
	\includegraphics[width=2.8in]{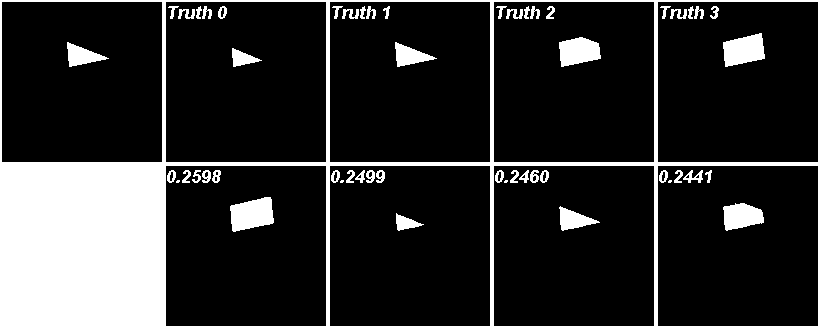}
    \includegraphics[width=2.8in]{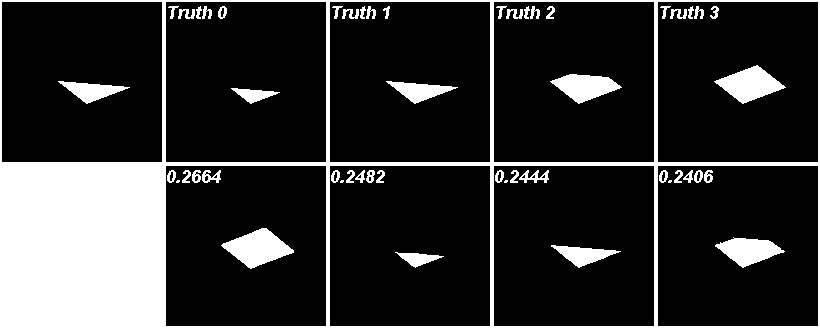}\\ 
	\caption{Results from our model on the shape reconstruction task. Results with the
predicted uncertainties ($>1e^{-5}$) are shown.}
	\label{shapes} 
\end{figure*}

Fig. \ref{shapes} shows the results of 8 more input shapes. Again, the optimal DMM accurately predicts the 4 labeled shapes and their associated probabilities. Overall, these results on the synthetic dataset validate the ability of the optimal DMM to not only generate the diverse set of shapes derived from the input triangle, but also provide reliable probability estimates for each predicted output. 

\subsection{Lung segmentation with pulmonary opacity}

We next test our proposed framework for lung segmentation of chest radiographs with pulmonary opacities. Lung segmentation of chest radiographs with pulmonary opacities is a challenging task. The presence of pulmonary opacities, such as infections, masses, or consolidations, can introduce significant uncertainty in accurately delineating the lung boundaries. The opacities can obscure the true lung margins, making it difficult to determine the exact extent of the lungs. Due to the ambiguity introduced by the pulmonary opacities, different radiologists or experts may provide varying lung segmentation labels for the same chest radiograph. There can be disagreement on where to draw the precise lung boundaries, especially in regions with opacity.

In this experiment, a collection of chest radiographs with pulmonary opacities are synthesized based on the lung X-ray dataset \cite{candemir2013lung,jaeger2013automatic} which comprises 640 data-label pairs for training and 63 pairs for testing with the size of $(180, 180)$. To simulate real-world scenarios, we introduce random intensity occlusions to intact the radiographs. The two corresponding segmentation masks labeled by two different experts are generated under different rules. The first-kind label is a segmentation mask from the original dataset, indicating the complete region of the lungs. The second kind differs from the original segmentation by randomly removing part of the opacity area to stimulate the doctors' empirical annotations of the lung masks. Using our proposed framework, our goal is to obtain an optimal DMM that fits the dataset with the labeled segmentation mask, as well as predicts the probability associated with each plausible segmentation result.

In the training process, just as the shape reconstruction task, we still randomly sample a chest radiograph with pulmonary opacities $x_i$ from the training set and randomly sample one of its two labels as $y_i^k$. Then the distribution of the label space consists of two modes with the same probability of $0.5$. 

\begin{figure*}[t]
	\centering
	\includegraphics[width=2.8in]{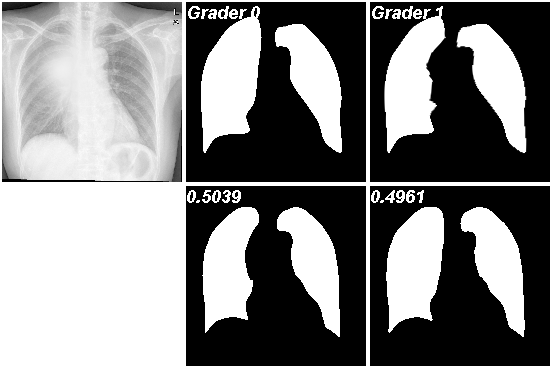}
    \includegraphics[width=2.8in]{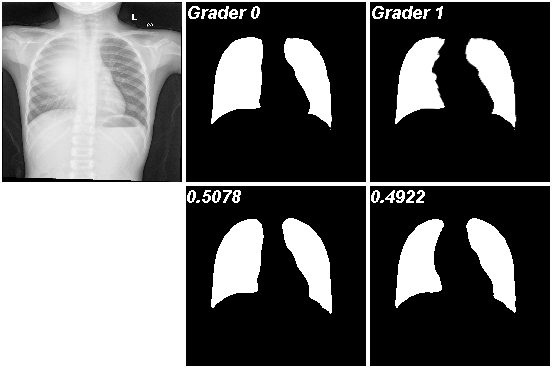}
	\caption{Results visualization for lung corrupt segmentation. The first row shows the input samples and their labels, and the next row shows the predictions from our method. The predicted probability for each output is annotated in the upper-left corner.}
	\label{lung} 
\end{figure*}

\begin{figure*}[t]
	\centering
	\includegraphics[width=2.8in]{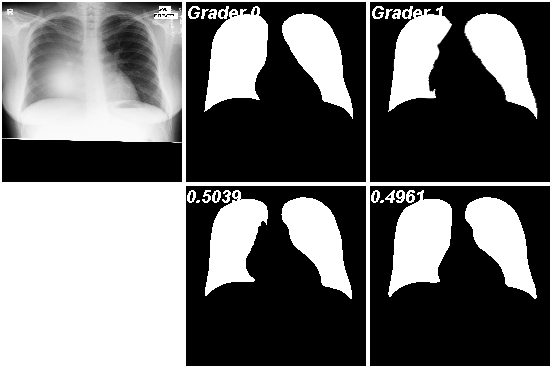}
    \includegraphics[width=2.8in]{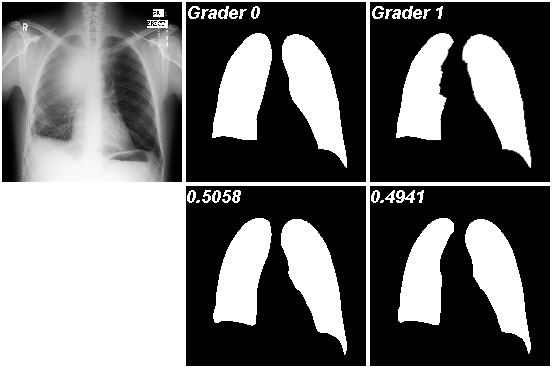}
	\includegraphics[width=2.8in]{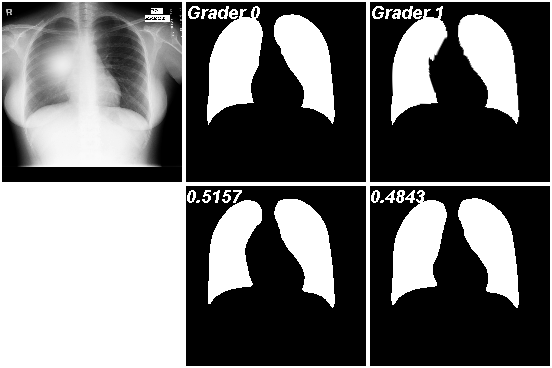}
    \includegraphics[width=2.8in]{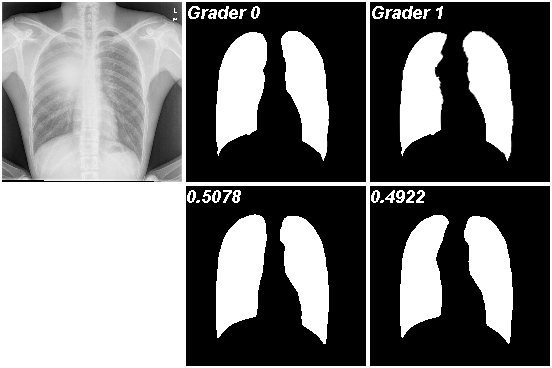}\\  
    \includegraphics[width=2.8in]{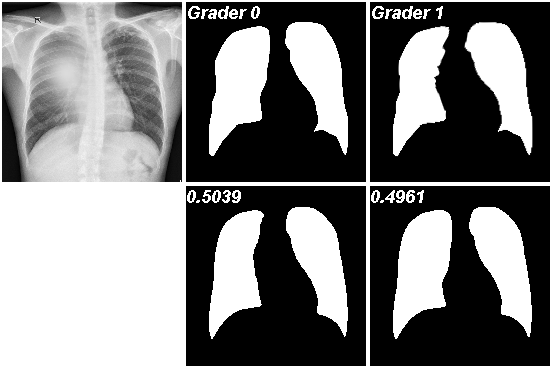}
    \includegraphics[width=2.8in]{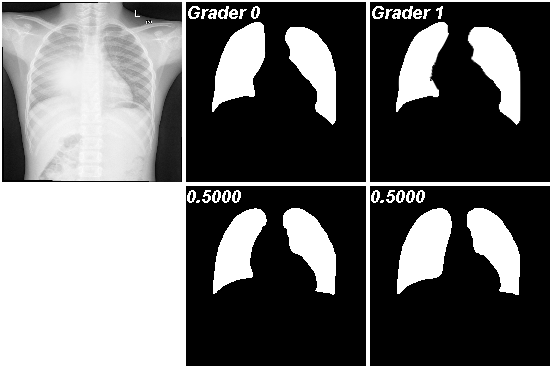}
	\includegraphics[width=2.8in]{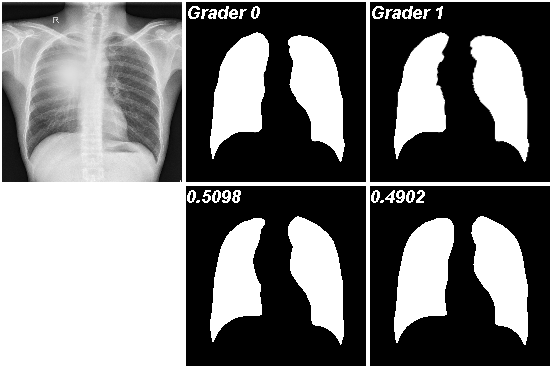}
    \includegraphics[width=2.8in]{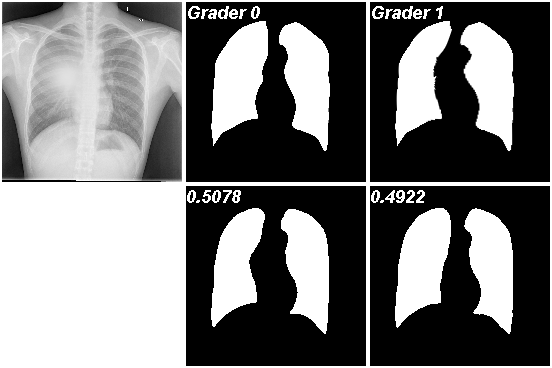}\\ 
    \includegraphics[width=2.8in]{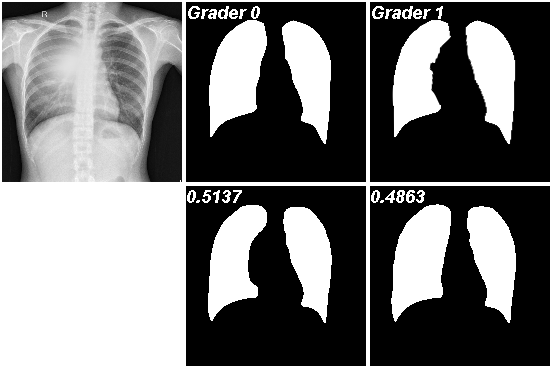}
    \includegraphics[width=2.8in]{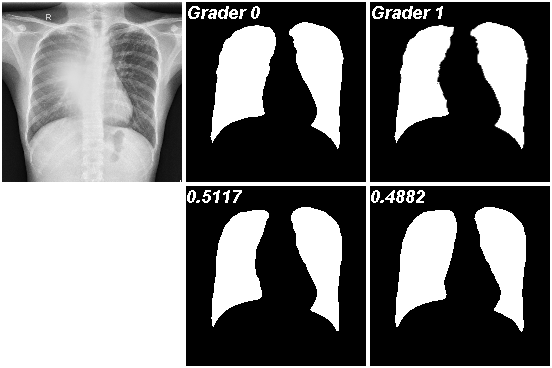}\\
	\caption{Results from our model on the lung corrupt segmentation task. Results with the
predicted uncertainties ($>1e^{-5}$) are shown.}
	\label{lungs} 
\end{figure*}

\begin{figure*}[t]
	\centering
	\includegraphics[width=2.8in]{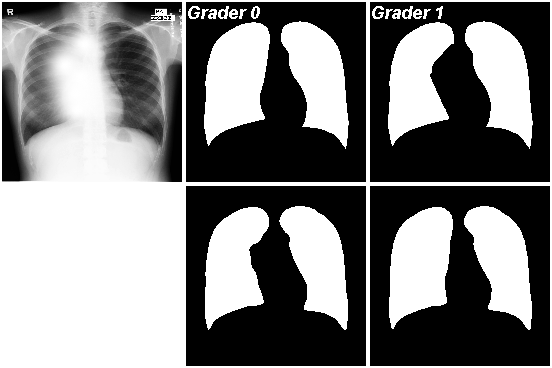}
    \includegraphics[width=2.8in]{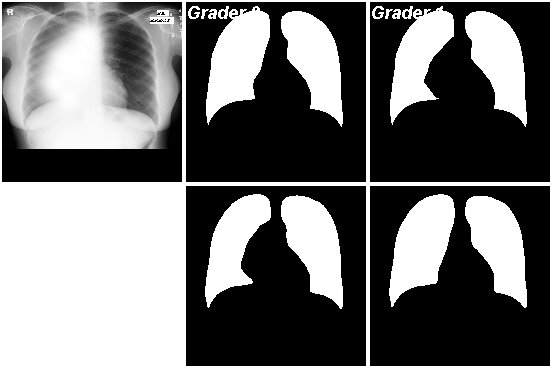}\\
    \includegraphics[width=2.8in]{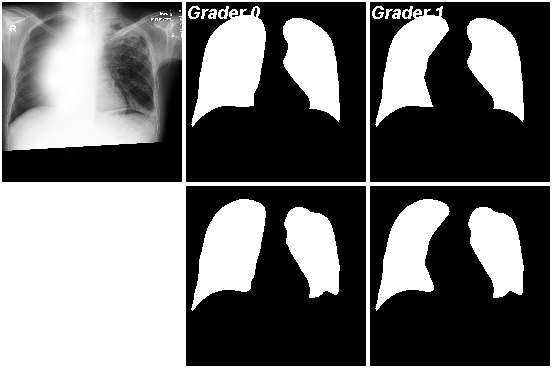}
    \includegraphics[width=2.8in]{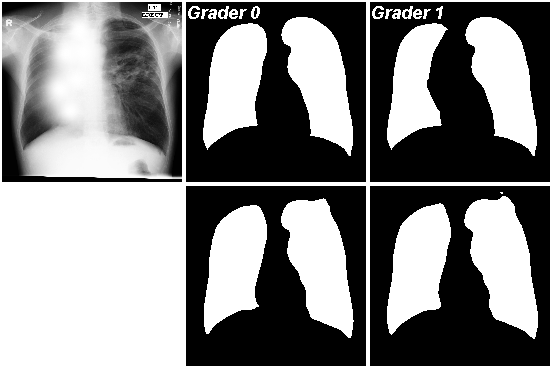}\\
	\includegraphics[width=2.8in]{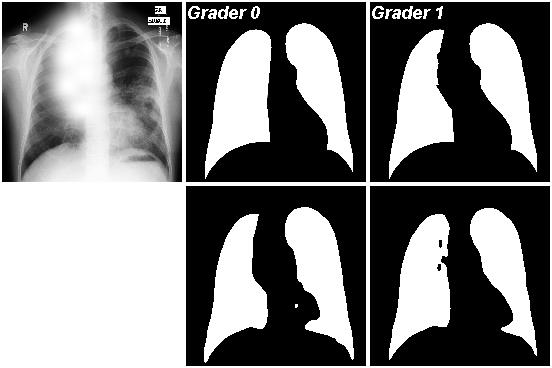}
    \includegraphics[width=2.8in]{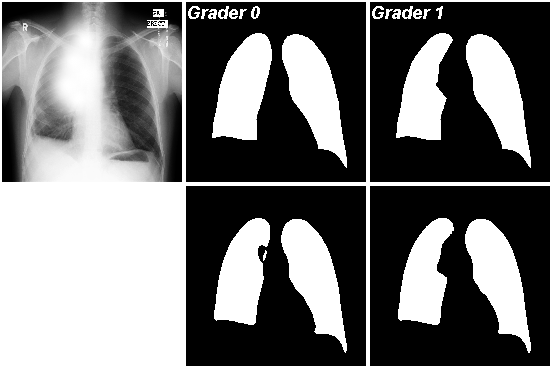}\\  
    \includegraphics[width=2.8in]{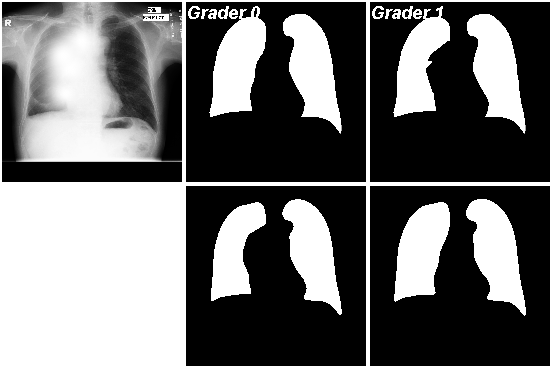}
    \includegraphics[width=2.8in]{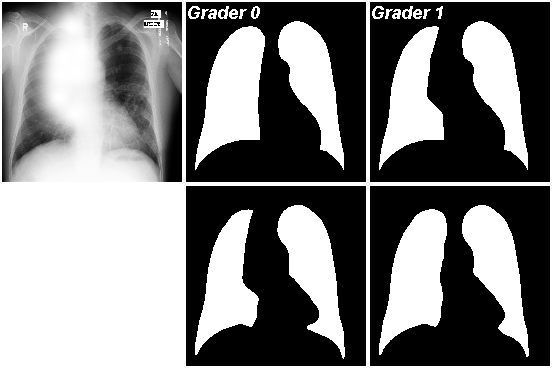}\\
	\includegraphics[width=2.8in]{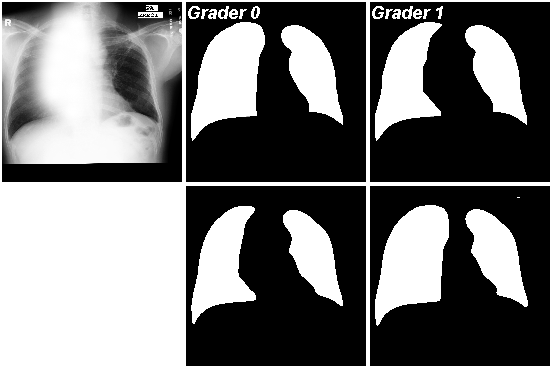}
    \includegraphics[width=2.8in]{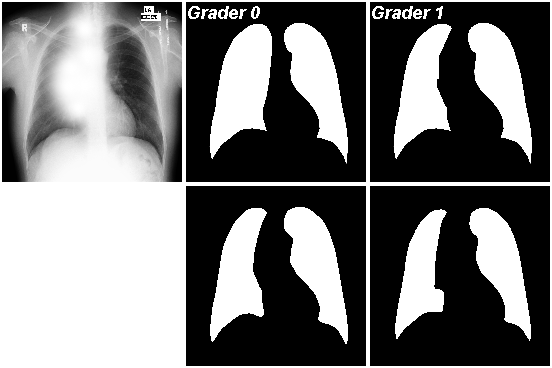}\\ 
	\caption{Results from Probabilistic U-Net on the lung corrupt segmentation task. 2 random sample results are shown.}
	\label{lung2_pu} 
\end{figure*}

\begin{figure*}[t]
	\centering
	\includegraphics[width=2.8in]{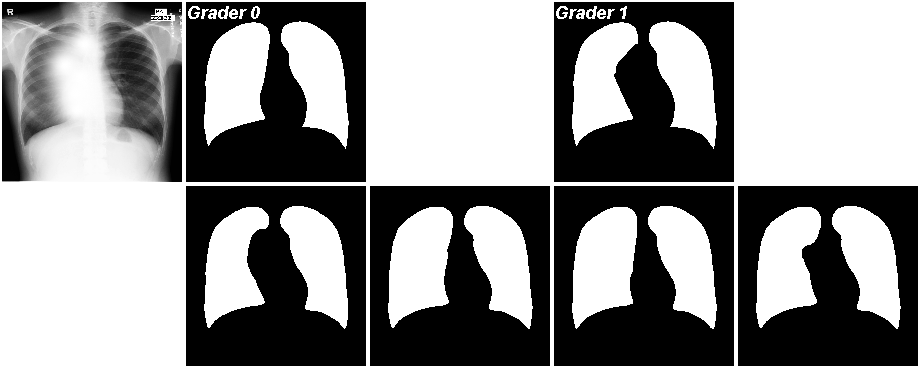}
    \includegraphics[width=2.8in]{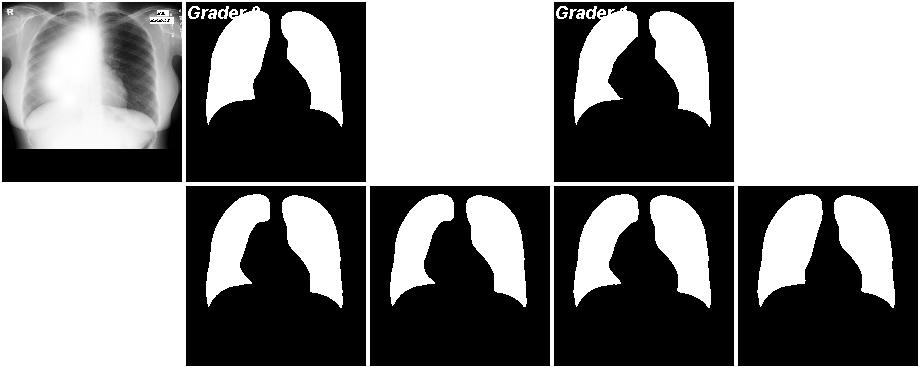}\\
    \includegraphics[width=2.8in]{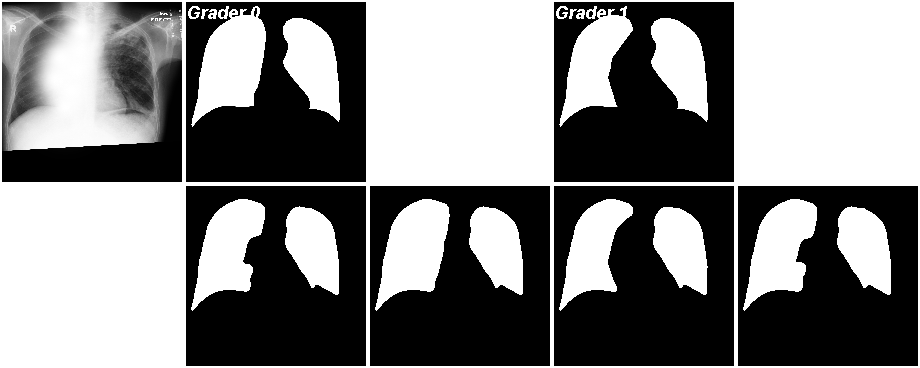}
    \includegraphics[width=2.8in]{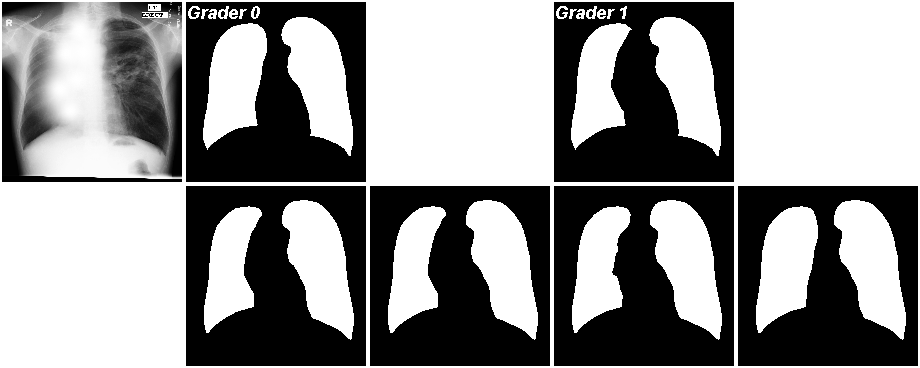}\\
	\includegraphics[width=2.8in]{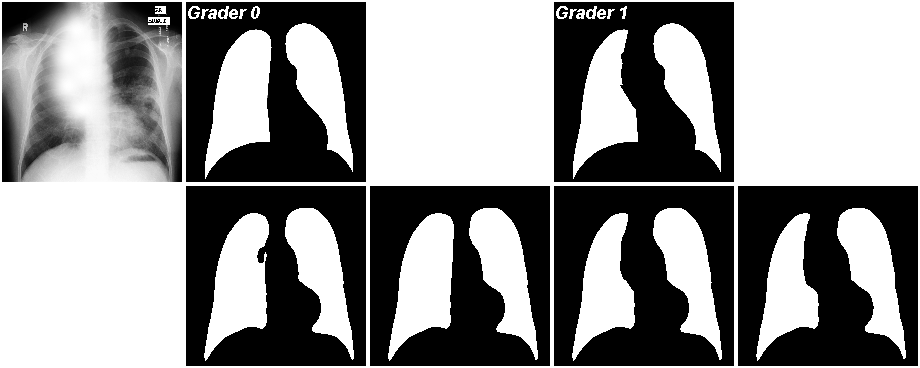}
    \includegraphics[width=2.8in]{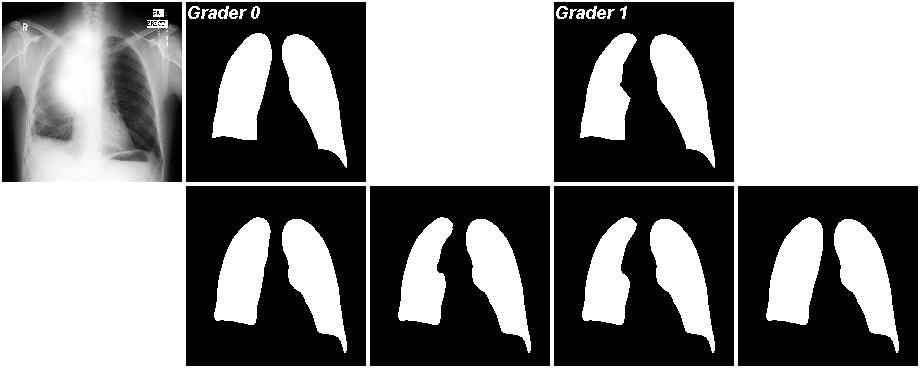}\\  
    \includegraphics[width=2.8in]{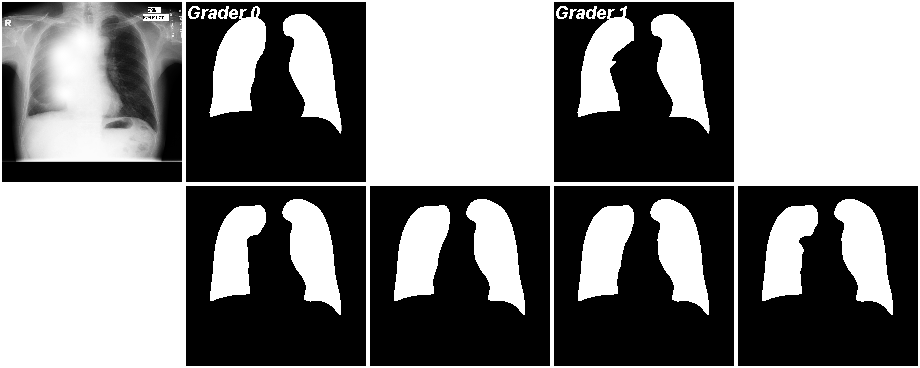}
    \includegraphics[width=2.8in]{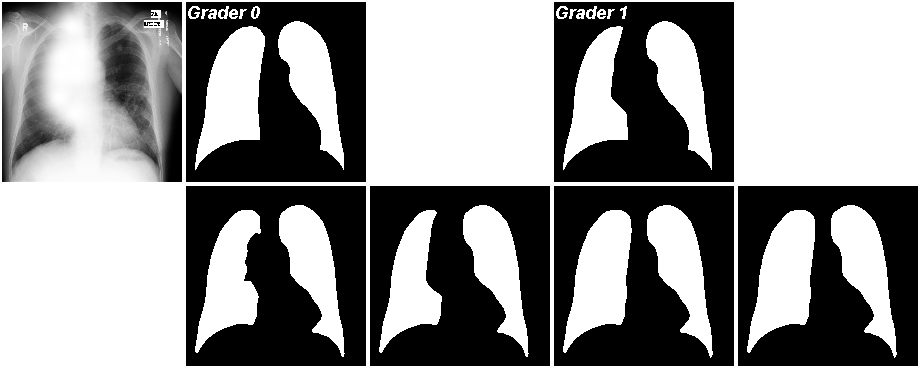}\\
	\includegraphics[width=2.8in]{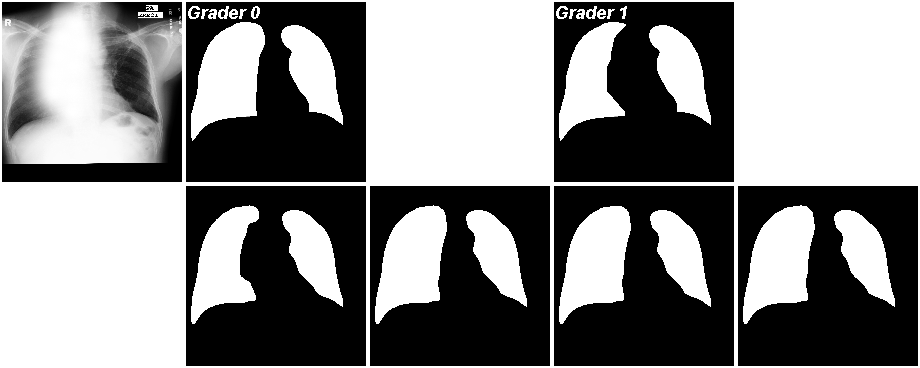}
    \includegraphics[width=2.8in]{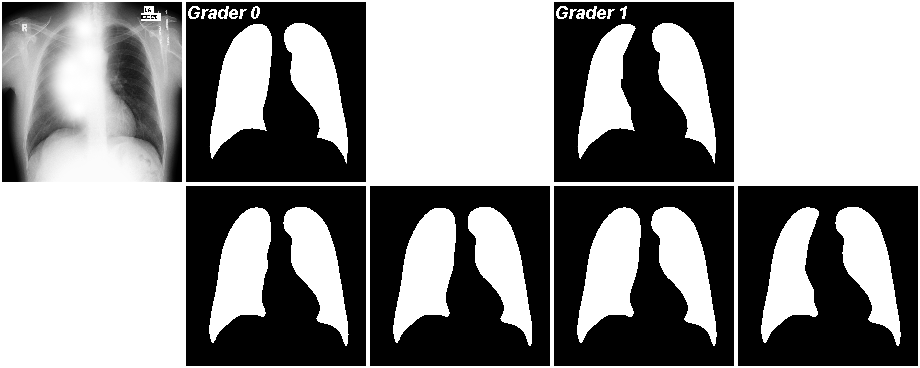}\\ 
	\caption{Results from Probabilistic U-Net on the lung corrupt segmentation task. 4 random sample results are shown.}
	\label{lung4_pu} 
\end{figure*}

\begin{figure*}[t]
	\centering
	\includegraphics[width=2.8in]{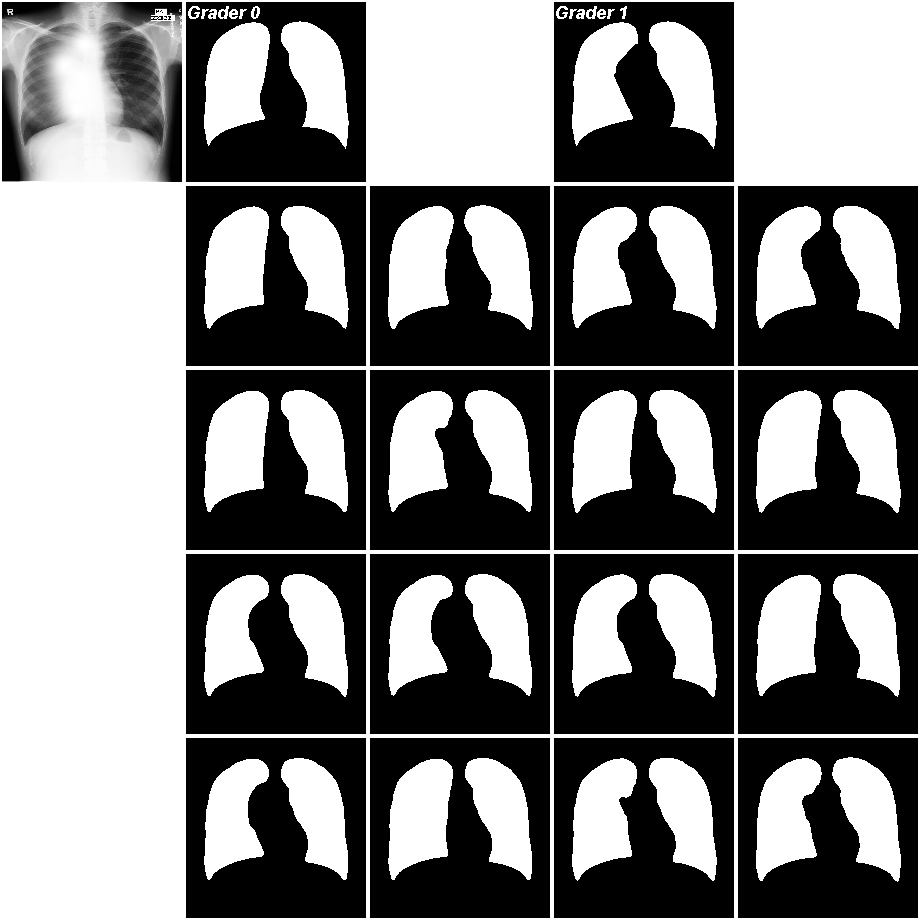}
    \includegraphics[width=2.8in]{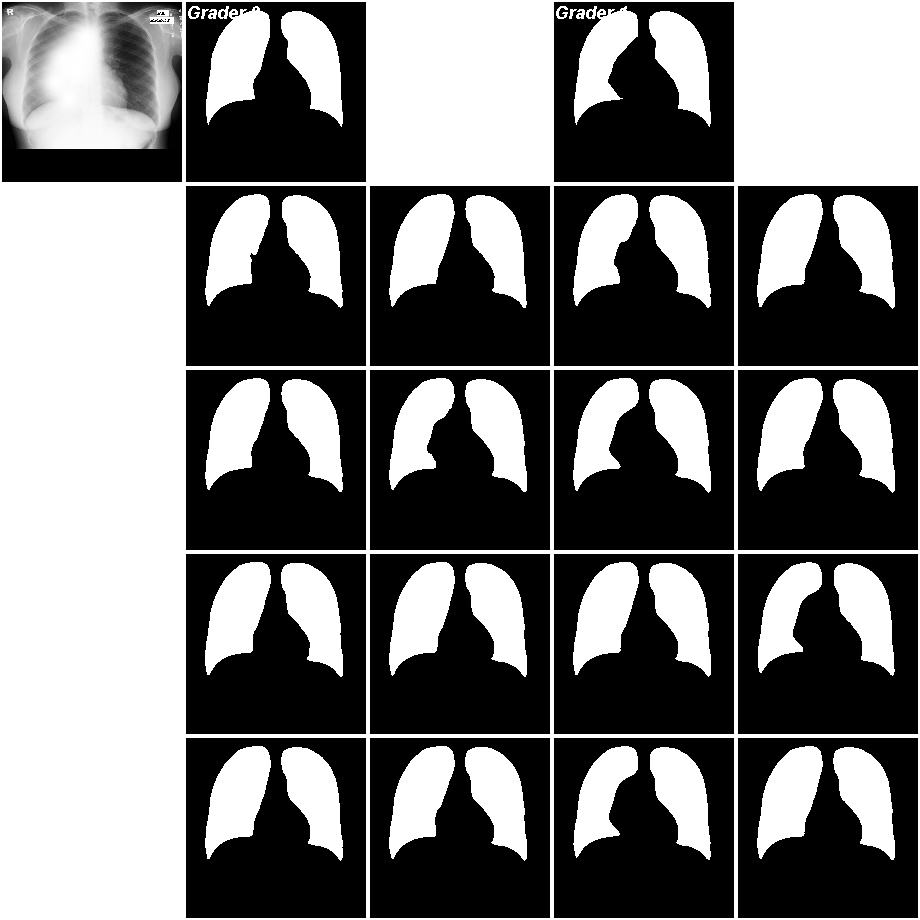}\\
    \includegraphics[width=2.8in]{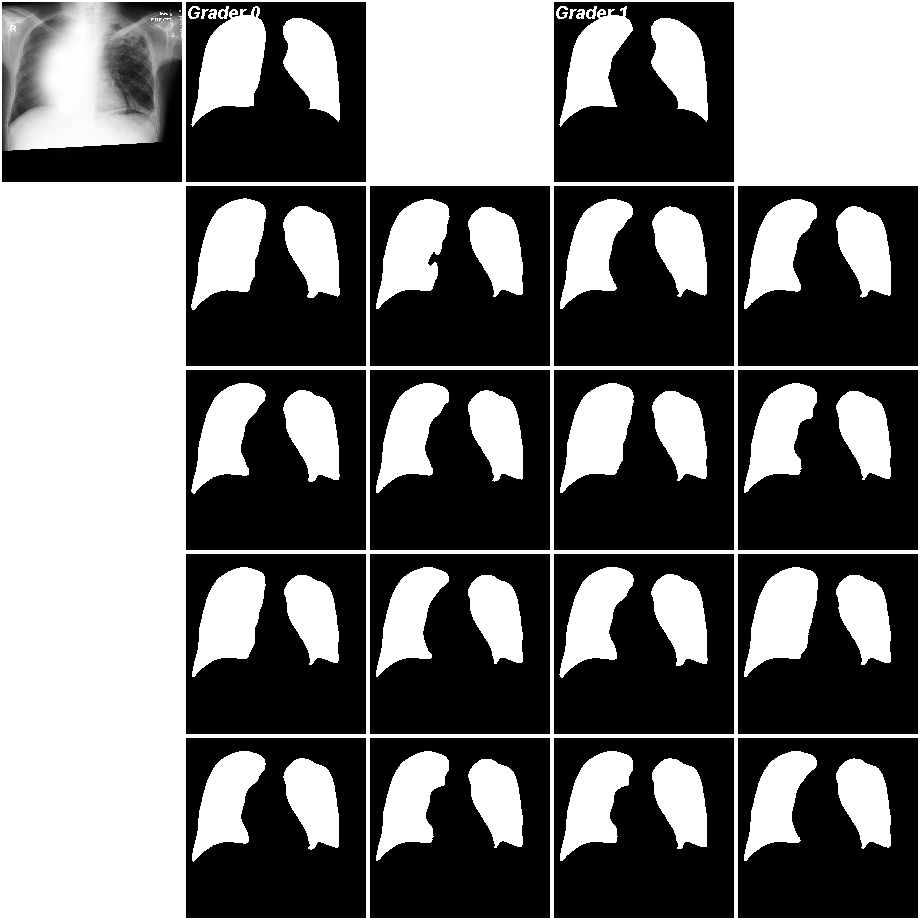}
    \includegraphics[width=2.8in]{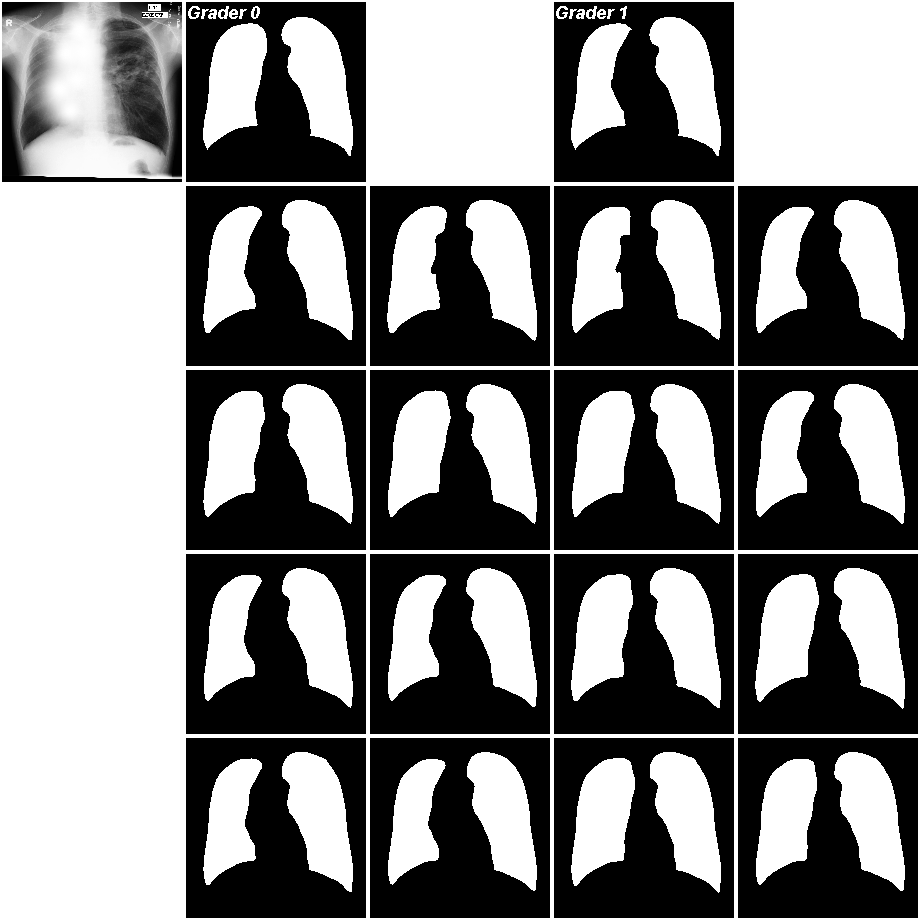}\\
	\includegraphics[width=2.8in]{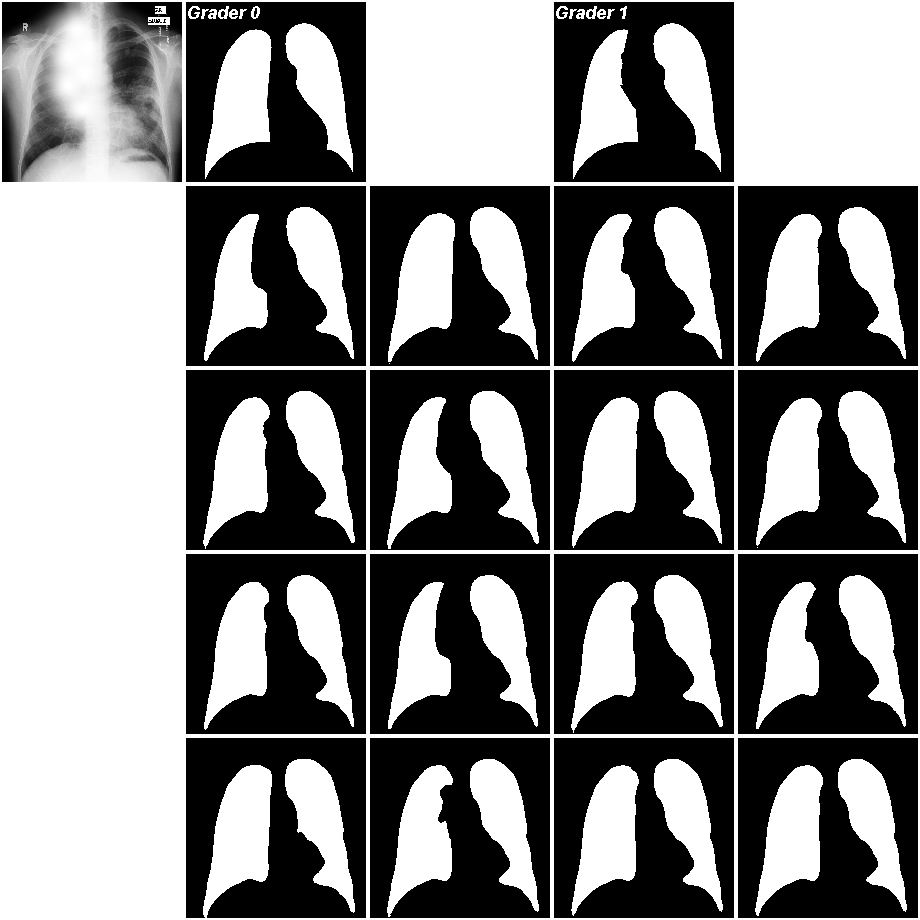}
    \includegraphics[width=2.8in]{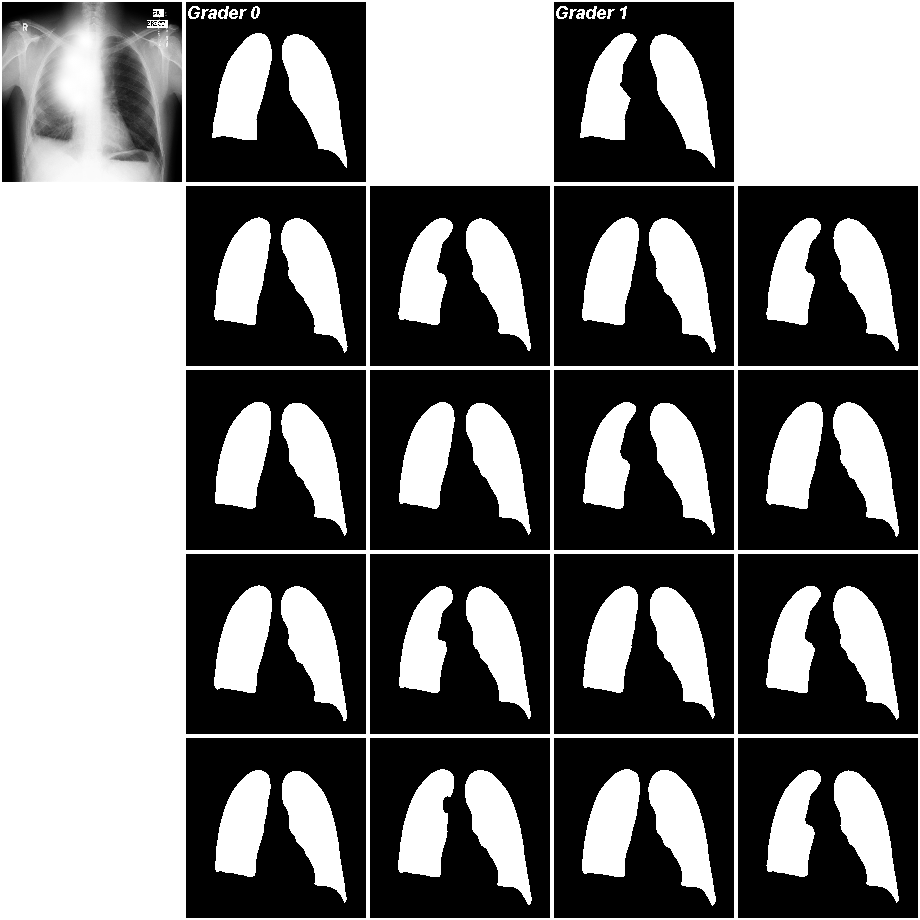}\\ 
	\caption{Results from Probabilistic U-Net on the lung corrupt segmentation task. 16 random sample results are shown.}
	\label{lung16_pu} 
\end{figure*}

To show the test results, we output predictions of chest radiograph from the testing set with explicit probability estimates ($P(\hat{y}_i^j|x_i) > 1e^{-5}$) annotated on the upper-left corner in the second row in Fig. \ref{lung}. More results are in Fig. \ref{lungs}. Obviously, the two outputs are very similar to the corresponding labels and have a probability close to $0.5$. We also trained the Probabilistic U-Net model on this dataset with its results shown in Fig. \ref{lung2_pu}, \ref{lung4_pu} and \ref{lung16_pu}.  Since it cannot directly predict the probability of the output, we respectively do $2$ samples, $4$ samples, and $16$ samples to estimate the results. For each mode's probability, we count the proportion of predictions similar to that label in all sampling results as the probability of that label. For example, in the results of doing $2$ samplings, if both results are similar to the first-kind label, then the probability of the first-kind label is 1, and the probability of the other label is 0. This means that it only predicts one mode. If the two results are similar to the two labels respectively, then the probability of both labels is 0.5, which is the correct prediction. The probabilities' distributions are shown in Fig. \ref{prob} and Table \ref{prob_mean}. It shows our model achieves excellent results in both segmentation and probabilistic prediction accuracy.

\begin{table}[]
\centering
\begin{tabular}{c|c|c|c|c}
\hline
    model&  Ours& Prob. U-net 2& Prob. U-net 4& Prob. U-net 16\\
\hline
   Grader 0 mean& 0.5040 & 0.4921 & 0.5238 &0.5496\\
\hline
   Grader 1 mean& 0.4960 & 0.5079 & 0.4762 &0.4504\\
\hline
     std& 0.0262 & 0.3391 & 0.2345 & 0.1482 \\
\hline
\end{tabular}
\caption{Mean and std values of predicted probabilities in Lung corrupt segmentation}
\label{prob_mean}
\end{table}

\subsection{Real applications}\label{exp_2} 
\begin{figure*}[t]
	\centering
	\includegraphics[width=2.8in]{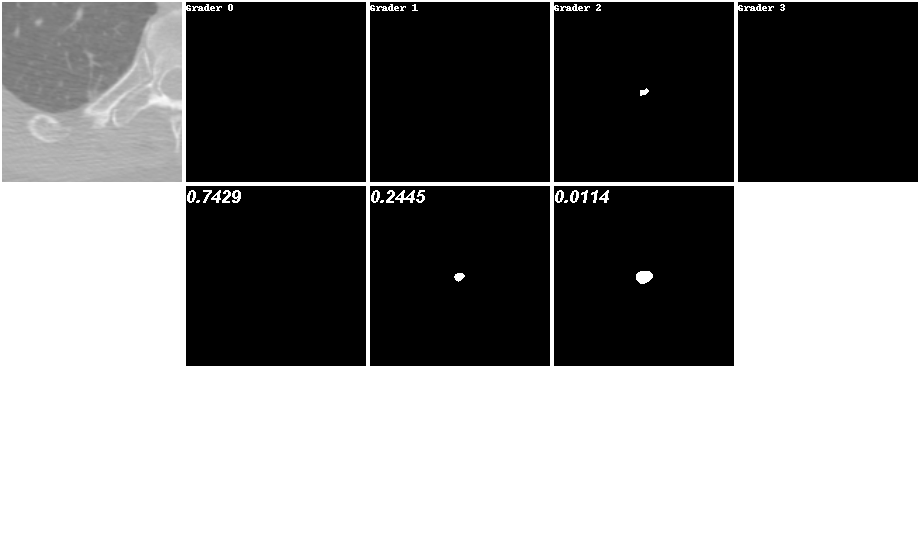}
    \includegraphics[width=2.8in]{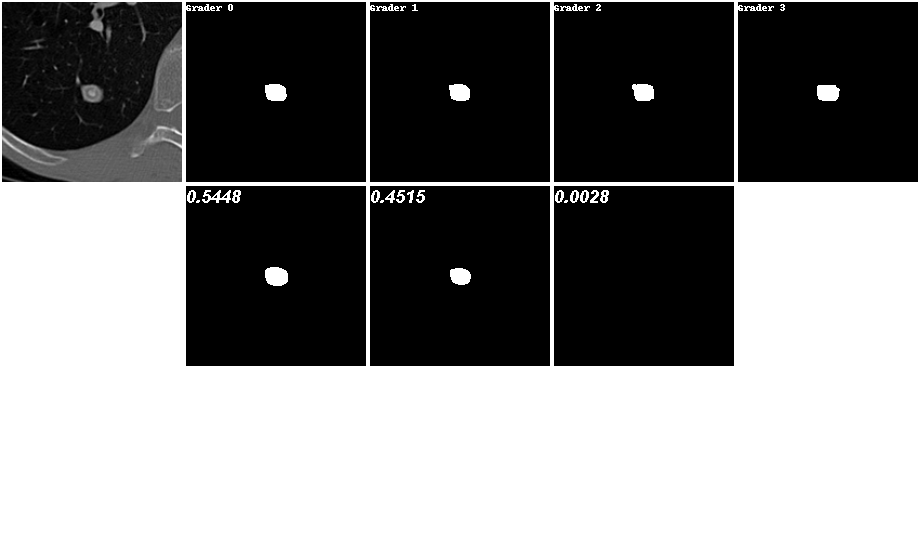}
	\caption{Results visualization for lung corrupt segmentation. The first row shows the input samples and their labels, and the next row shows all the predictions from our method with probabilities larger than $1e^{-5}$. The predicted probability for each output is annotated in the upper-left corner.}
	\label{LIDC} 
\end{figure*}

To evaluate the performance of our model on more complex real-world datasets, we work on the lesion segmentation task of ambiguous lung CT scans.  In this experiment, the LIDC-IDRI dataset is provided by \cite{armato2011lung,clark2013cancer}, which contains 1018 lung CT scans from 1010 patients. It is an unorganized multi-modal distributed dataset. Each scan has four (out of twelve) medical experts labeling it a mask. And experts independently judged the location and shape of existing lesions based on their respective knowledge and experience. This does not mean that the label set is a balanced 4-modal discrete distribution. In actual situations, for the same scan, the masks given by experts may be the same, but when it comes to another scan, the masks they give will become different. The first rows of Fig. \ref{LIDC} that are sampled from the testing set can prove this. They even have objections to whether a lesion exists in the scan. Our task is to learn these imbalanced multi-modal distributions and make accurate uncertainty estimates while outputting possible outcomes. 

\begin{figure*}[t]
	\centering
	\includegraphics[width=2.8in]{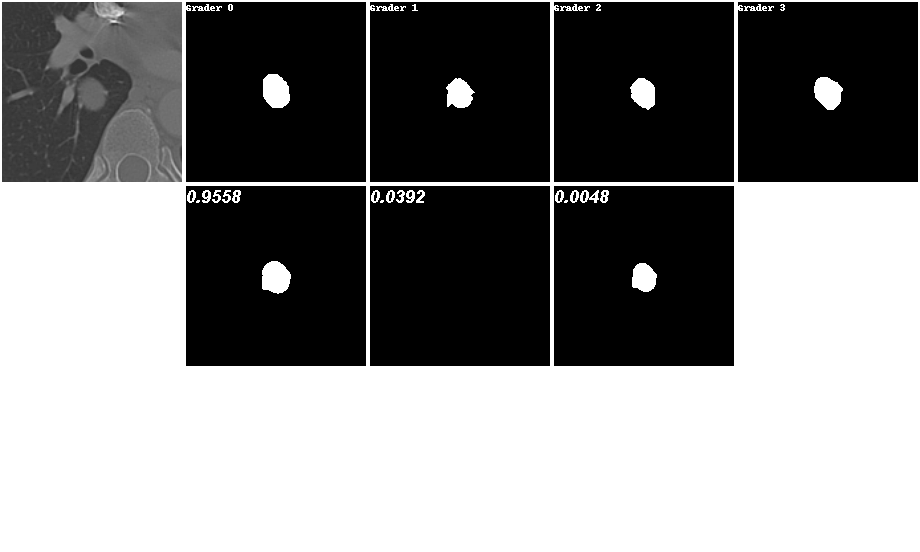}
    \includegraphics[width=2.8in]{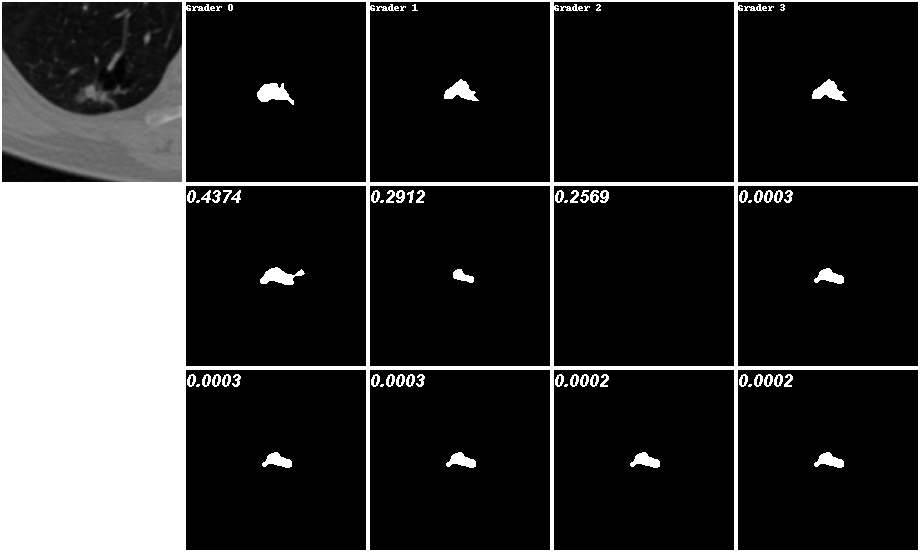}\\
	\includegraphics[width=2.8in]{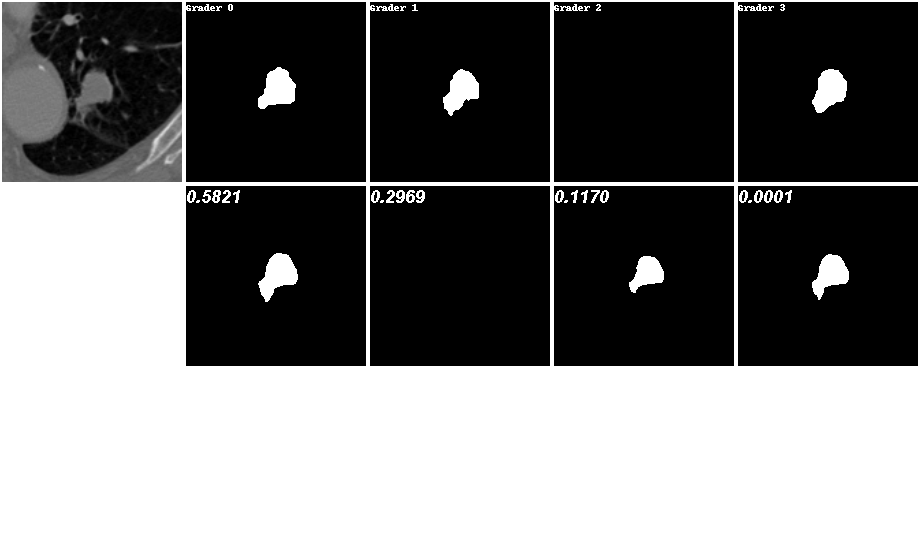}
    \includegraphics[width=2.8in]{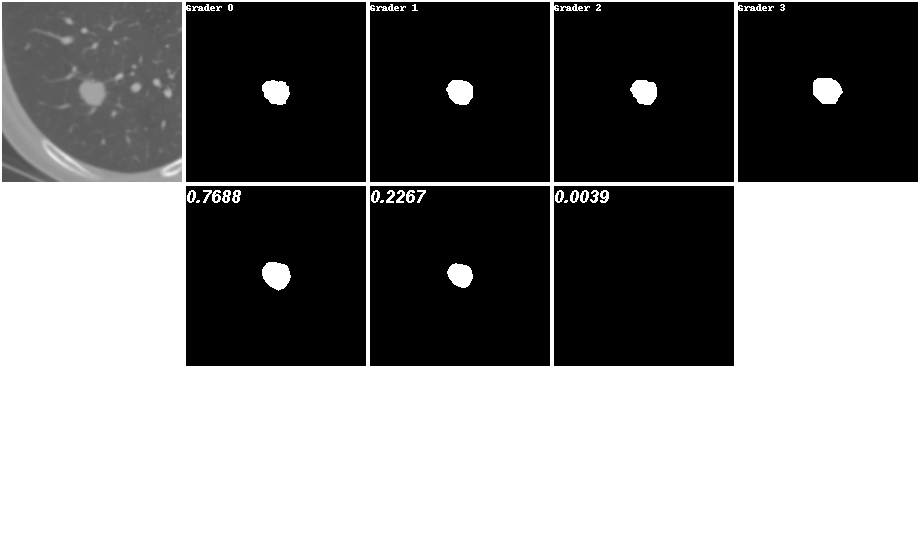}\\  
    \includegraphics[width=2.8in]{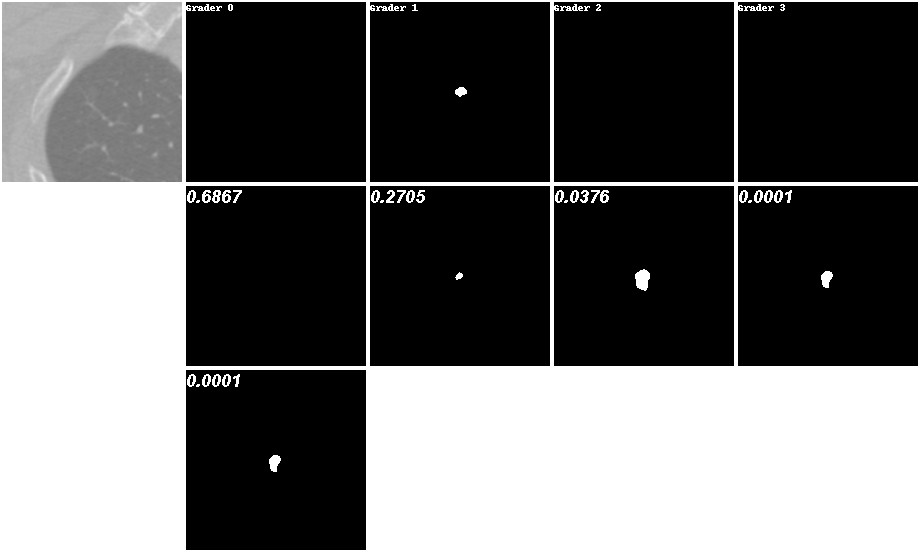}
    \includegraphics[width=2.8in]{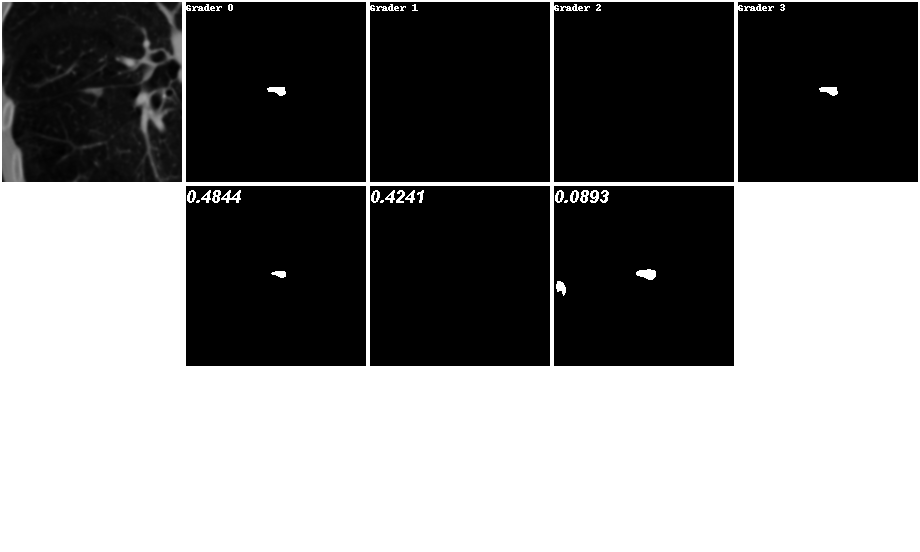}\\
	\includegraphics[width=2.8in]{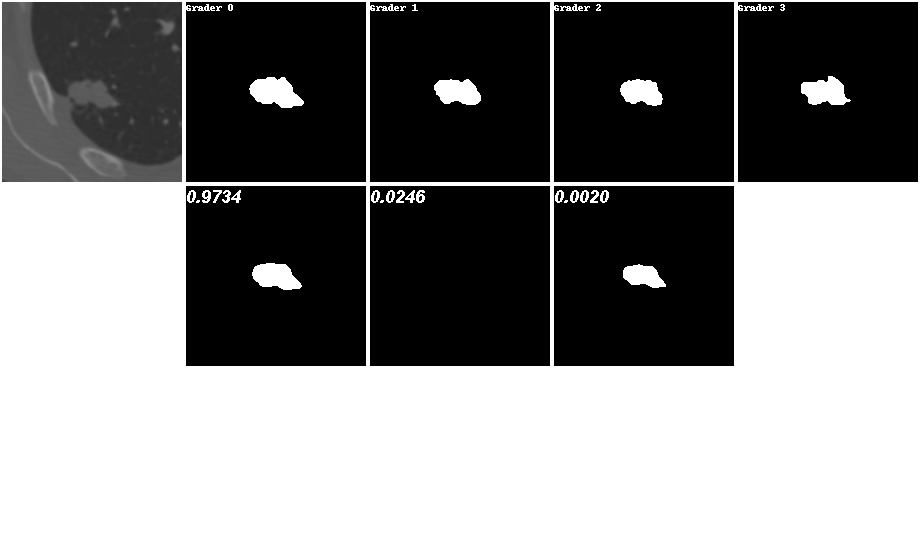}
    \includegraphics[width=2.8in]{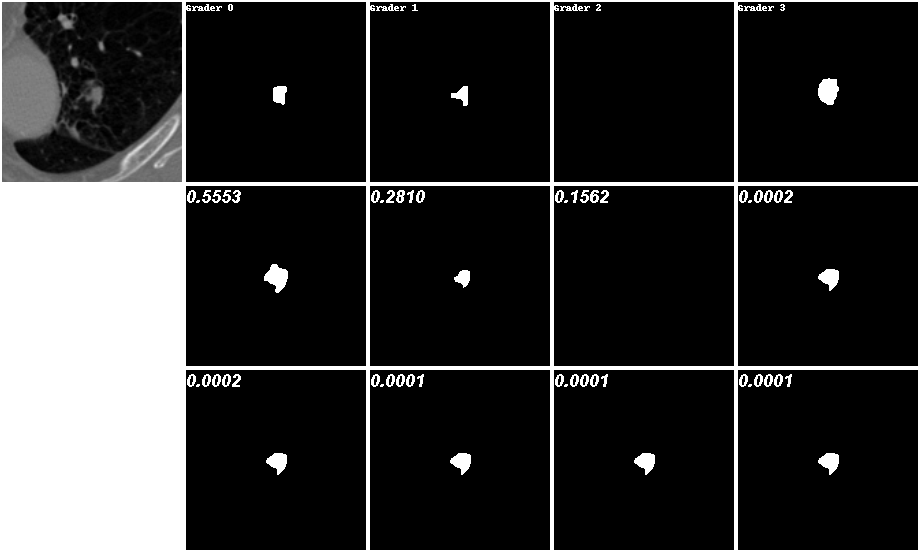}\\ 
    \includegraphics[width=2.8in]{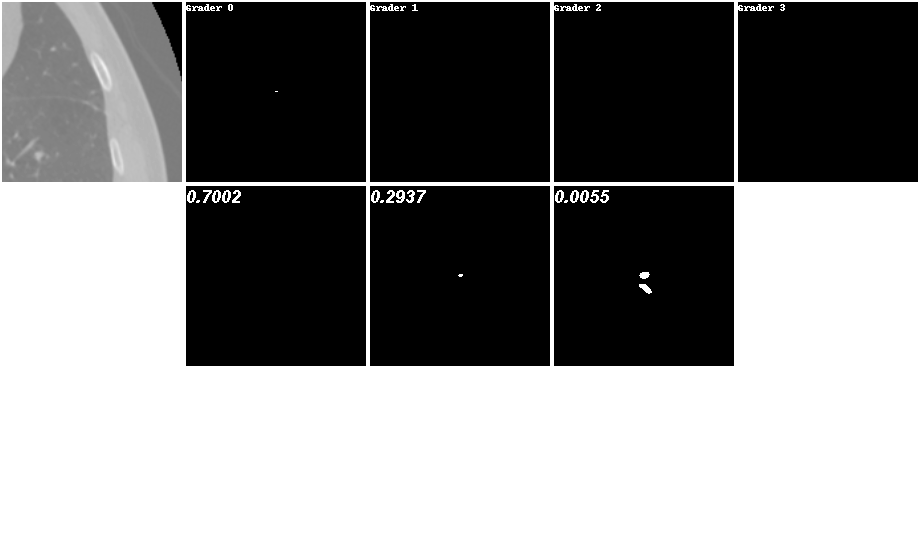}
    \includegraphics[width=2.8in]{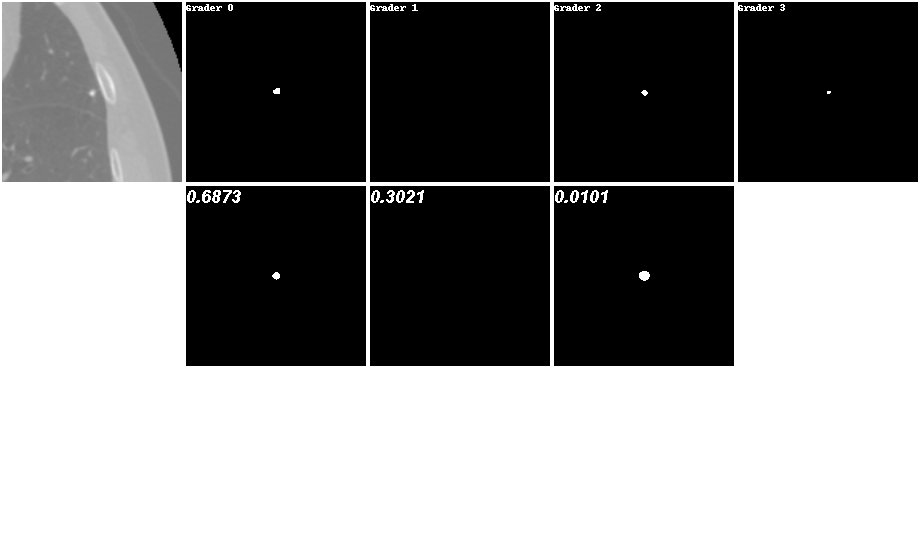}\\ 
	\caption{Results from our model on the LIDC-IDRI segmentation task. Results with the
predicted uncertainties ($>1e^{-5}$) are shown.}
	\label{LIDCs} 
\end{figure*}

During the training process, we randomly sample a CT scan $x_i$ from the training set and one of its four segmentations as $y_i^k$. Fig. \ref{LIDC} shows some examples from the testing set with high precision. The first row is the lesion scan and its four labels. The last two rows (We have reserved eight positions for the results in the picture.) are our top predictions, where the probability (larger than $1e^{-5}$) associated with each prediction is annotated on the upper-left corner. More results are shown in Fig. \ref{LIDCs}. It's obvious that our method effectively captures the uncertainty present in the segmentation labels, as evidenced by significant probability scores. The probabilities of predictions are almost equal to the probabilities of the true distribution. The no lesion inference on the left is especially approximated at $0.75$, and on the right is close to $0$. 
       
Unlike synthetic datasets, there is no ground truth distribution for the LIDC-IDRI dataset. In order to evaluate the model performance, we adopt the generalized energy distance metric $D^2_{\text{GED}}$ found in \cite{bellemare2017cramer,szekely2013energy}, which only access the samples from the distributions that models induce. It is used to evaluate the performance of the Probabilistic U-net. However, the Probabilistic U-net model cannot generate dynamic result sets and corresponding uncertainty estimates. Thus, we rewrite this metric according to the type of output we get as follows:
given the label set $Y_{x} \subset Y$ and the prediction set $S_{x} \subset Y$ which are corresponding to the data $x$, the general energy distance metric of them is
$$
D_{\mathrm{GED}}^2\left(Y_{x}, S_{x}\right)=2 \sum_{y \in Y_{x}} \sum_{s \in S_{x}} p_{s} p_{y} d(y, s)-\sum_{y \in Y_{x}} \sum_{y^{\prime} \in Y_{x}} p_{y} p_{y^{\prime}} d\left(y, y^{\prime}\right)-\sum_{s \in S_{x}} \sum_{s^{\prime} \in S_{x}} p_{s} p_{s^{\prime}} d\left(s, s^{\prime}\right),
$$
where $d(y, s)=1-\operatorname{IoU}(y, s)$ is the metric for evaluating the similarity of masks. $p_{s}$ is the probability prediction for the output $s$ and $p_{y}$ is the probability for the ground truth $y$. In case the ground truth is not available like LIDC-IDRI, we use $p_{y}=\frac{1}{\left|Y_{x}\right|}$, where $\left|Y_{x}\right|$ denotes the cardinality of $Y_{x}$. In particular, $p_{y} = \frac{1}{4}$ on the LIDC-IDRI task. For the Probabilistic U-Net, there are no probability predictions for each output. We set $p_{s} = \frac{1}{\left|S_{x}\right|}=\frac{1}{n}$ if we have $n$ samples from the model.

\begin{figure*}
\centering
\begin{subfigure}[t]{0.48\textwidth}
   \includegraphics[width=\linewidth]{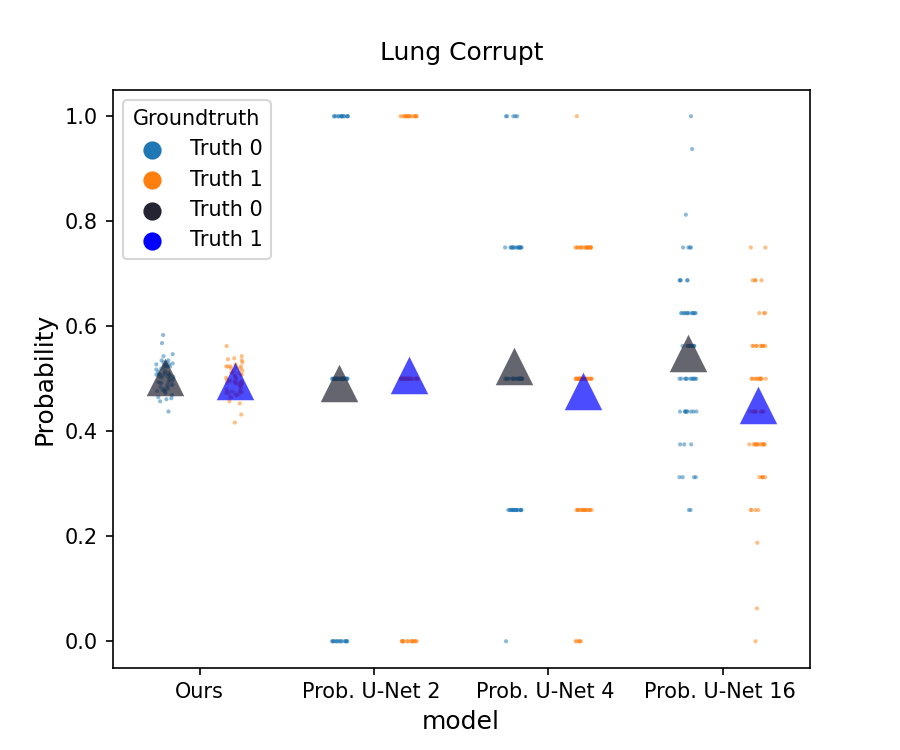}
   \caption{}
   \label{prob}
\end{subfigure}   
\begin{subfigure}[t]{0.48\textwidth}
   \includegraphics[width=\linewidth]{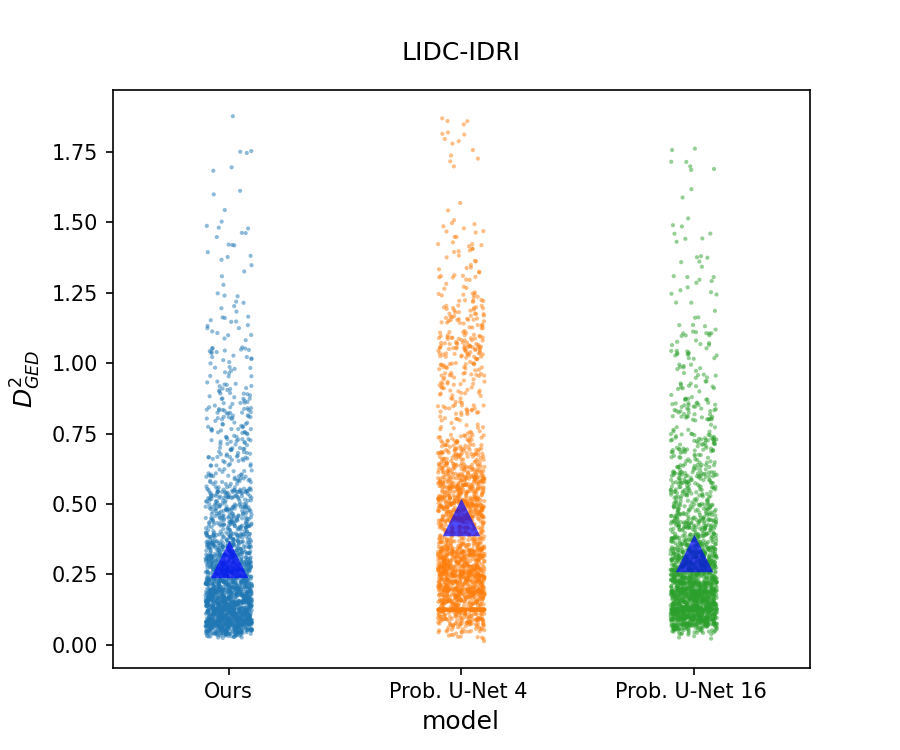}
   \caption{}
   \label{ged} 
\end{subfigure} 
\caption{Quantitative comparison. The small dots represent the quantities for test samples, and the small triangles represent the means of quantities. The numbers behind Prob. U-net are that of sample predictions. (a) shows our model produces an accurate uncertainty estimate for each mode. Probabilistic U-Net uses conventional Gaussian latent parametrization so that we can only sample results of a fixed number, e.g., 2, 4, 16, ... and then count the corporation of similar predictions as shown in the last three couples. However, the results are far less accurate than ours on the lung corrupt dataset. (b) shows the $D^2_{\text{GED}}$ values of test data on LIDC-IDRI segmentation task. Our model makes the state-of-the-art performance } 
\label{Quan} 
\end{figure*}

\begin{figure*}[t]
	\centering
	\includegraphics[width=2.8in]{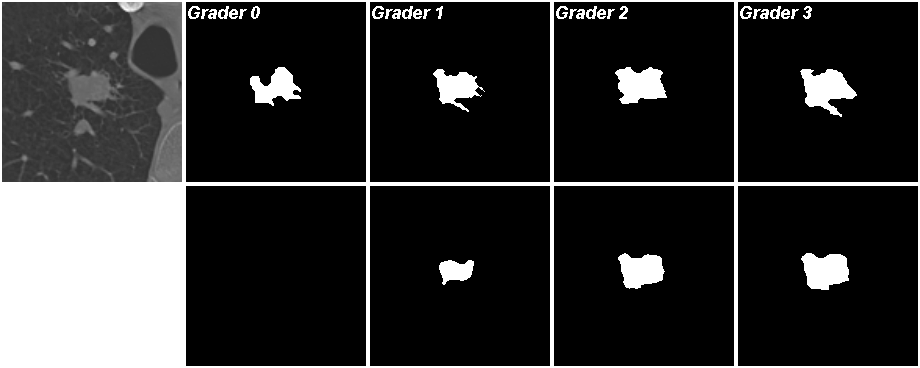}
    \includegraphics[width=2.8in]{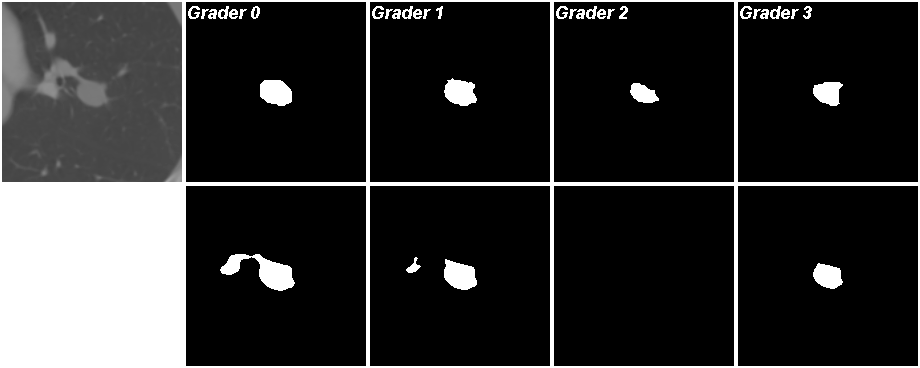}\\
	\includegraphics[width=2.8in]{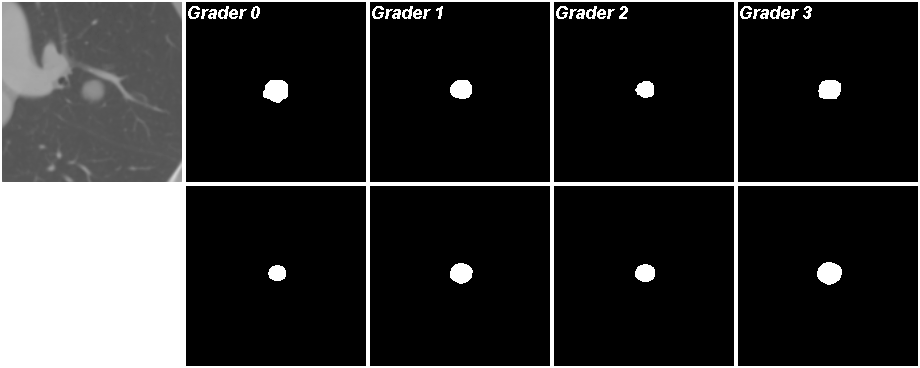}
    \includegraphics[width=2.8in]{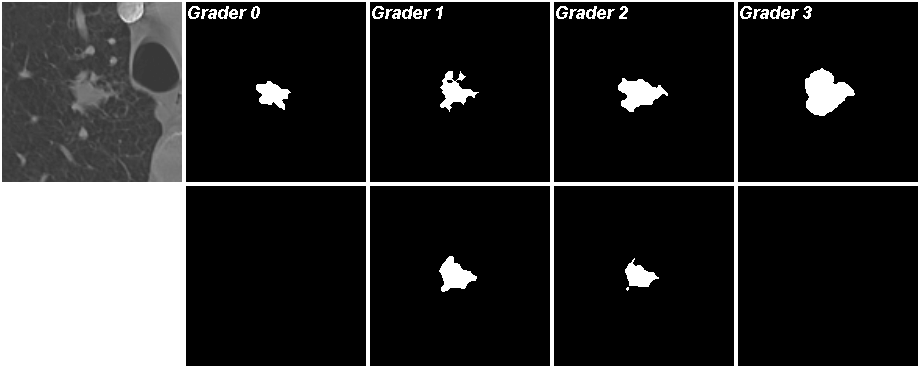}\\  
    \includegraphics[width=2.8in]{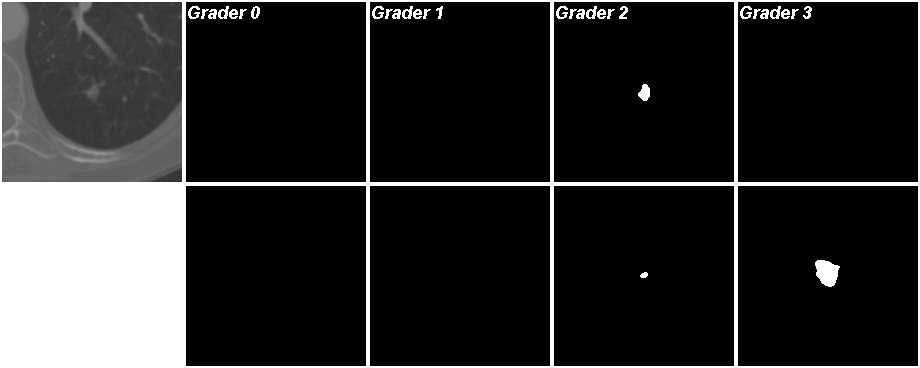}
    \includegraphics[width=2.8in]{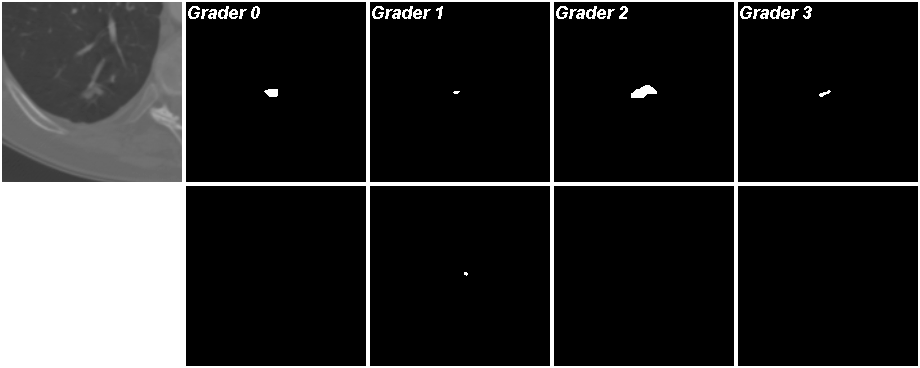}\\
	\includegraphics[width=2.8in]{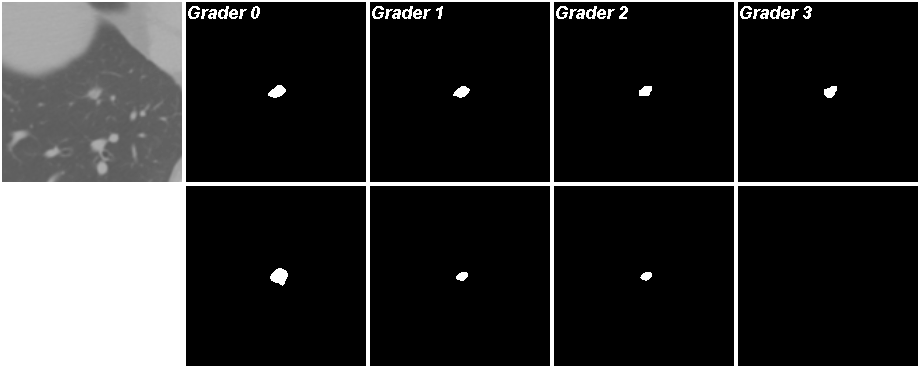}
    \includegraphics[width=2.8in]{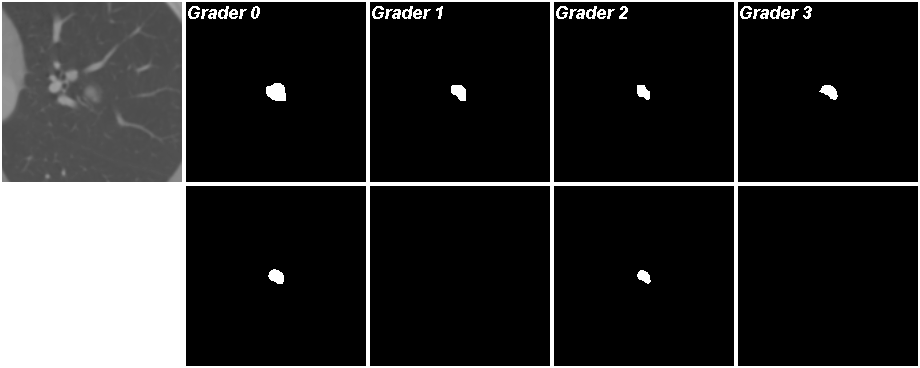}\\ 
    \includegraphics[width=2.8in]{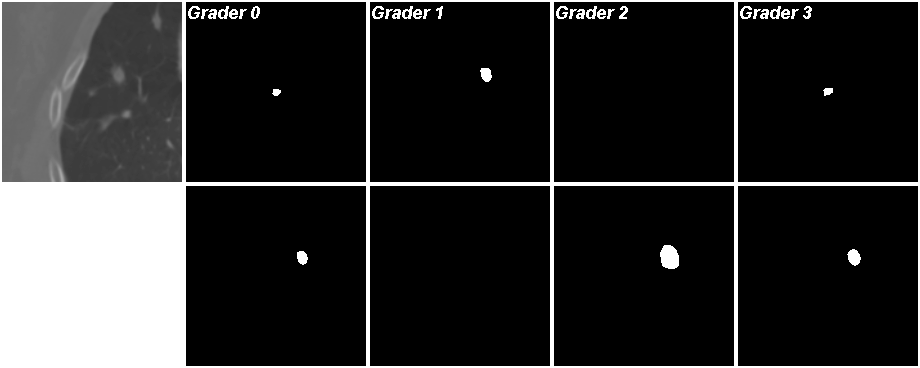}
    \includegraphics[width=2.8in]{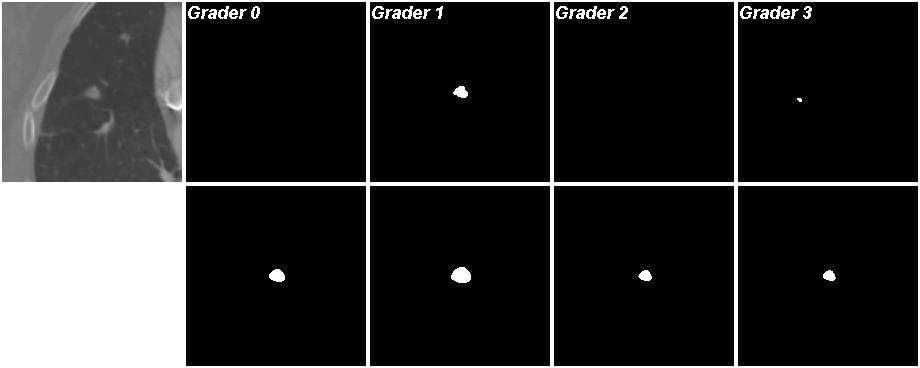}\\ 
	\caption{Results from Probabilistic U-Net on the LIDC-IDRI segmentation task. 4 random sample results are shown.}
	\label{LIDC4_pu} 
\end{figure*}

\begin{figure*}[t]
	\centering
	\includegraphics[width=2.8in]{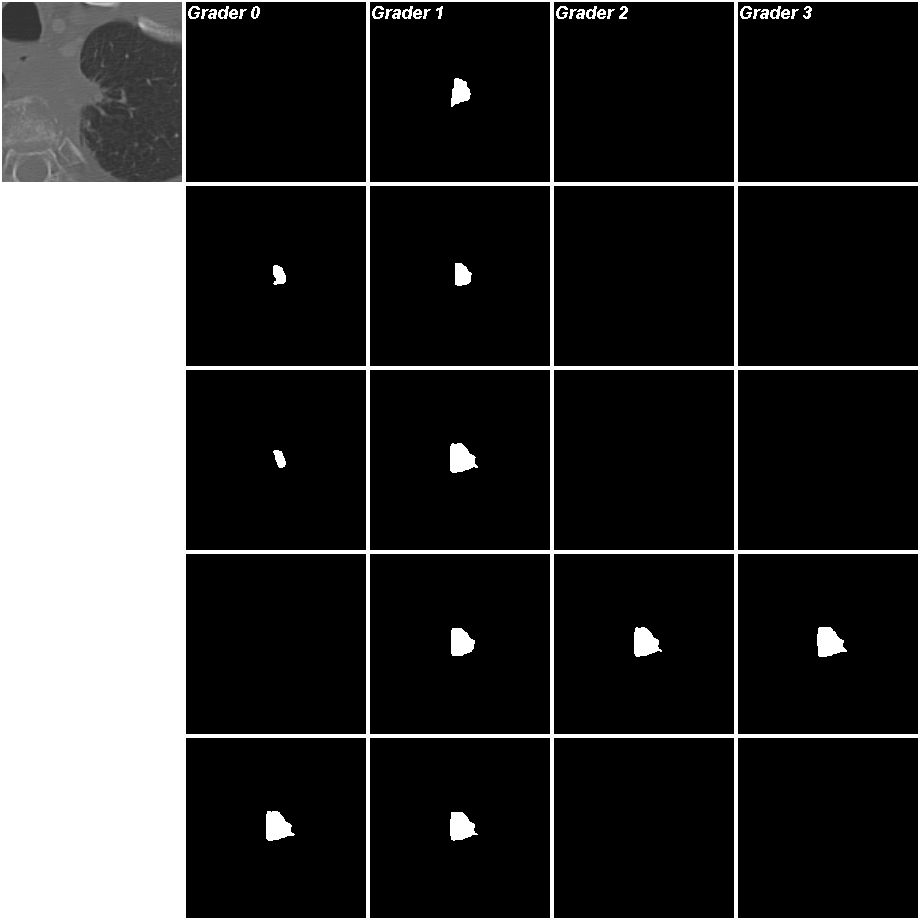}
    \includegraphics[width=2.8in]{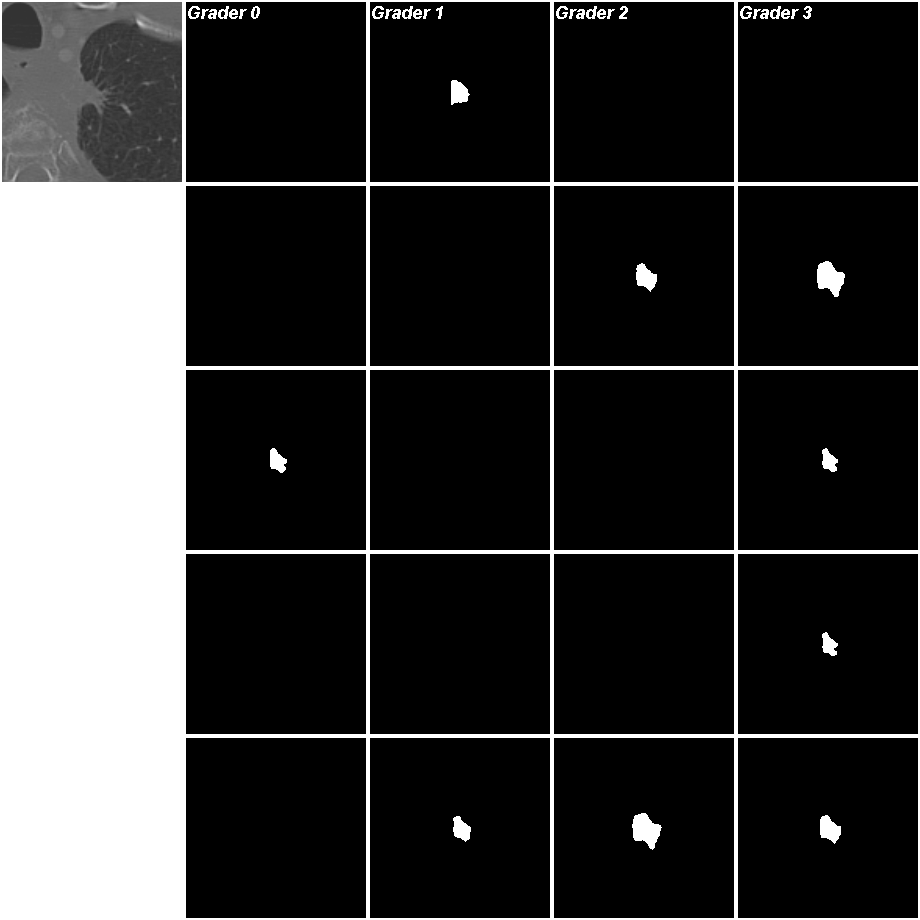}\\
	\includegraphics[width=2.8in]{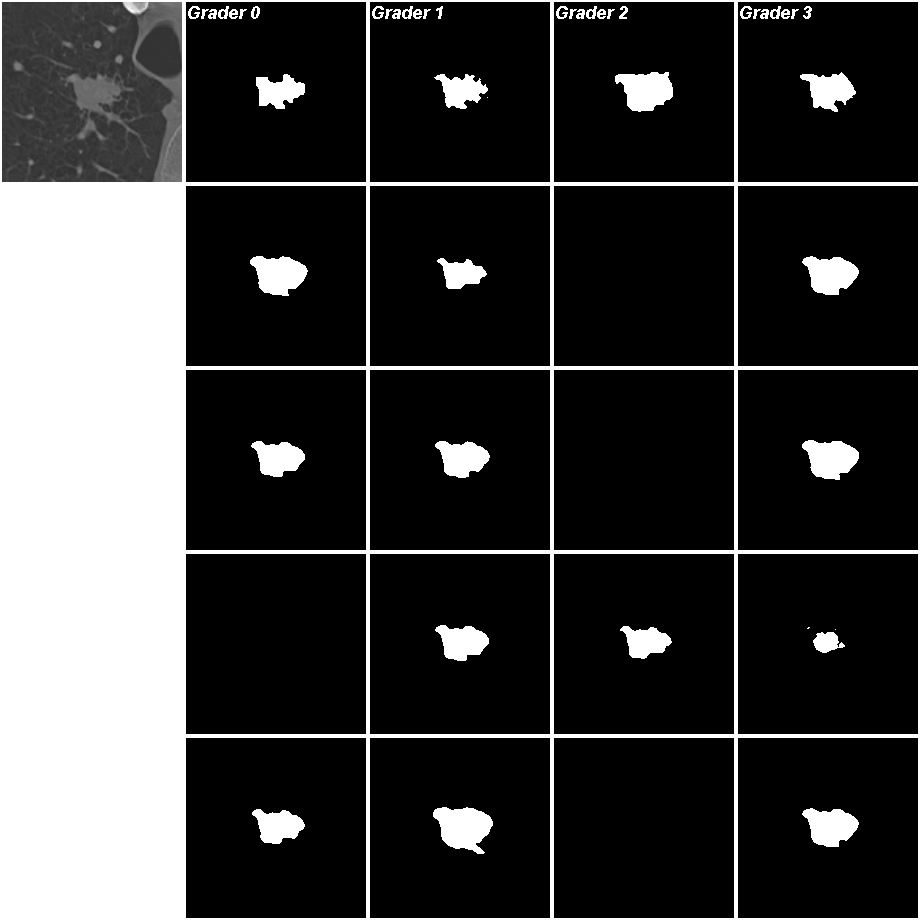}
    \includegraphics[width=2.8in]{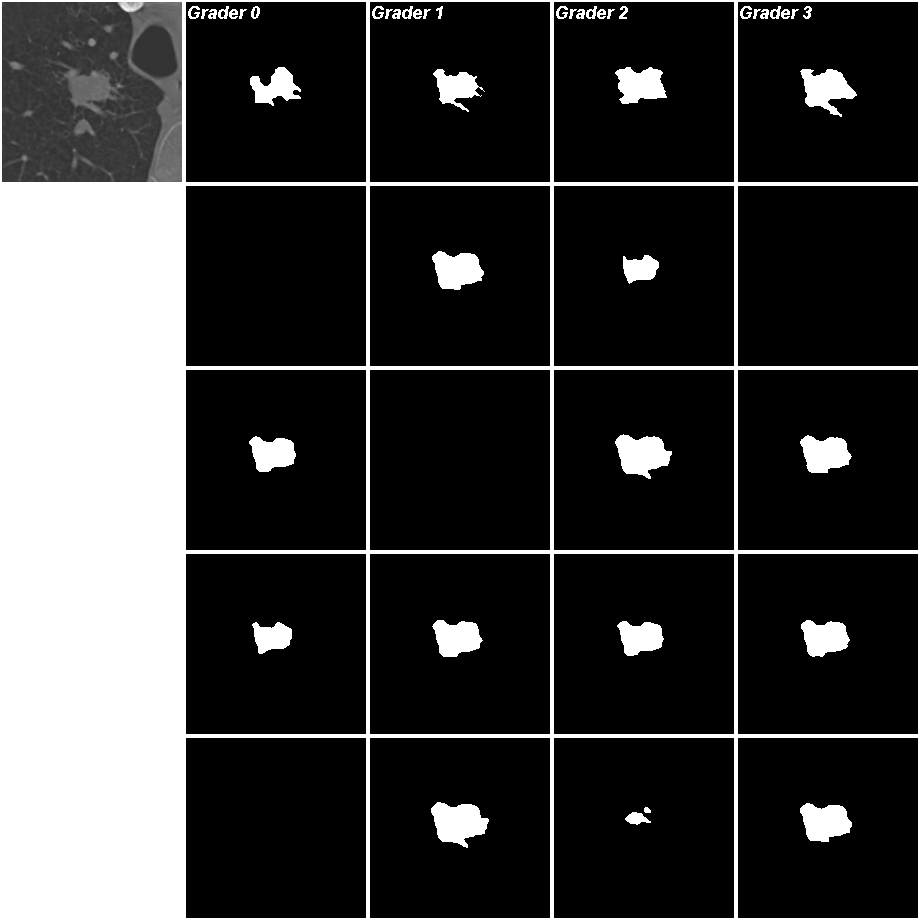}\\ 
    \includegraphics[width=2.8in]{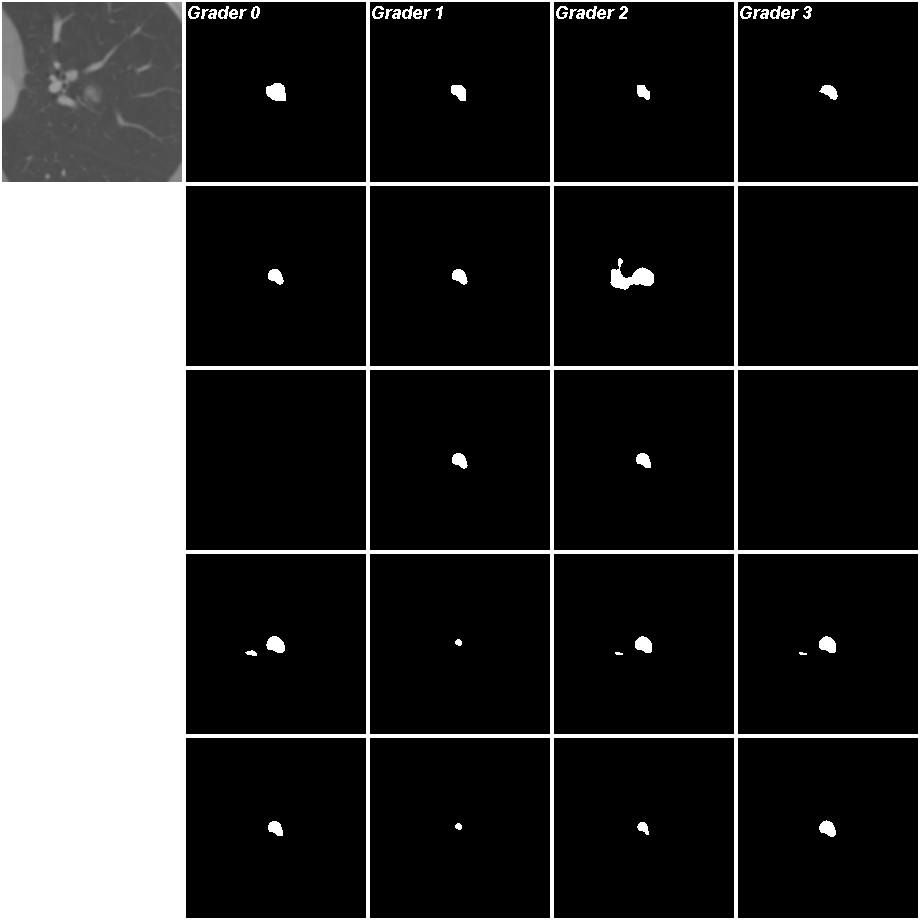}
    \includegraphics[width=2.8in]{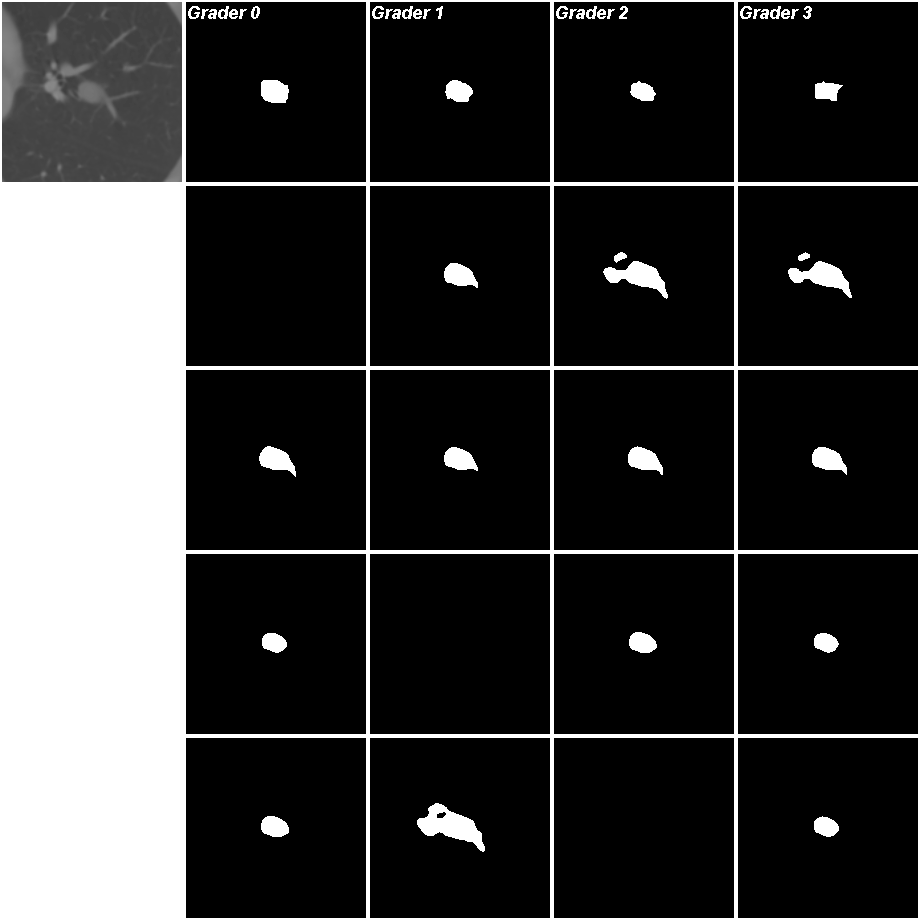}\\ 
	\caption{Results from Probabilistic U-Net on the LIDC-IDRI segmentation task. 16 random sample results are shown.}
	\label{LIDC16_pu} 
\end{figure*}

The quantity results of our model and Probabilistic U-net model are shown in Fig. \ref{ged} and Table \ref{ged_mean}. Lower values demonstrate the performance superiority of our model. More test predictions from Probabilistic U-Net are shown in Fig. \ref{LIDC4_pu} and Fig. \ref{LIDC16_pu}.

\begin{table}[]
\centering
\begin{tabular}{c|c|c|c}
\hline
         model&  Ours&  Prob. U-net 4& Prob. U-net 16\\
\hline
         mean&  0.3058&  0.4552& 0.3253\\
\hline
         std&  0.2761&  0.3375& 0.2743\\
\hline
\end{tabular}
\caption{Generalized Energy Distance in LIDC}
\label{ged_mean}
\end{table}

\subsection{ABLATION ANALYSIS}In this subsection, we explore some tricks applied in the model and some terms in the loss function in order to verify how they promote the model. 

\begin{figure*}[t]
	\centering
	\includegraphics[width=2.8in]{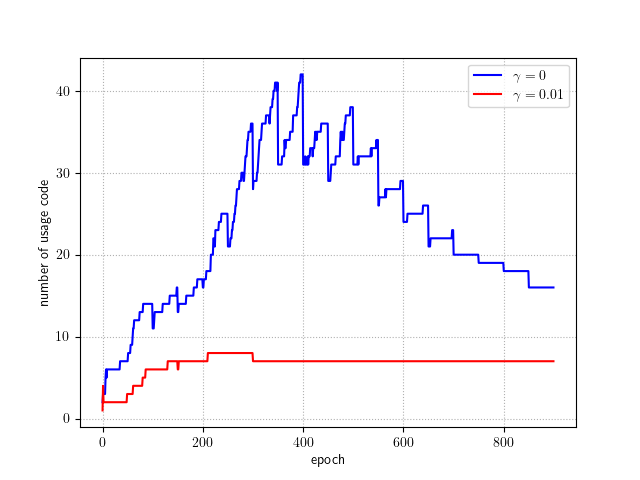}
    \includegraphics[width=2.8in]{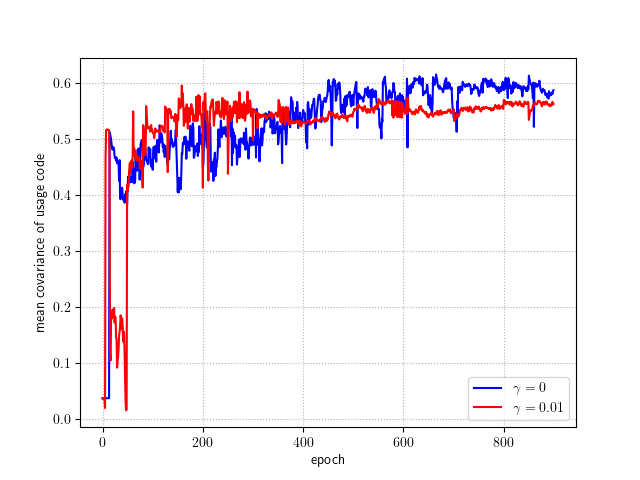}
	\caption{Ablations analysis. The left shows the number of usage codes in each epoch during training, and the right shows the mean of the pairwise inner product of these codes.}
	\label{cov} 
\end{figure*}

To test the performance of covariance loss, we train our model on the LIDC dataset by setting the weight of covariance loss as $\gamma=0$ and $\gamma=0.01$, respectively. Given the same initial codebook, we count the number of codes called in each epoch and their average pairwise inner products in the training process. The results are shown in \ref{cov}. We can see that covariance can help us separate the codes so that the features of the data can be concentrated on codes with a small number. On the contrary, without the balance of covariance loss, the frequency of use of codes will fluctuate significantly, and the codes will be more similar to each other.

\begin{figure*}[t]
	\centering
	\includegraphics[width=2.8in]{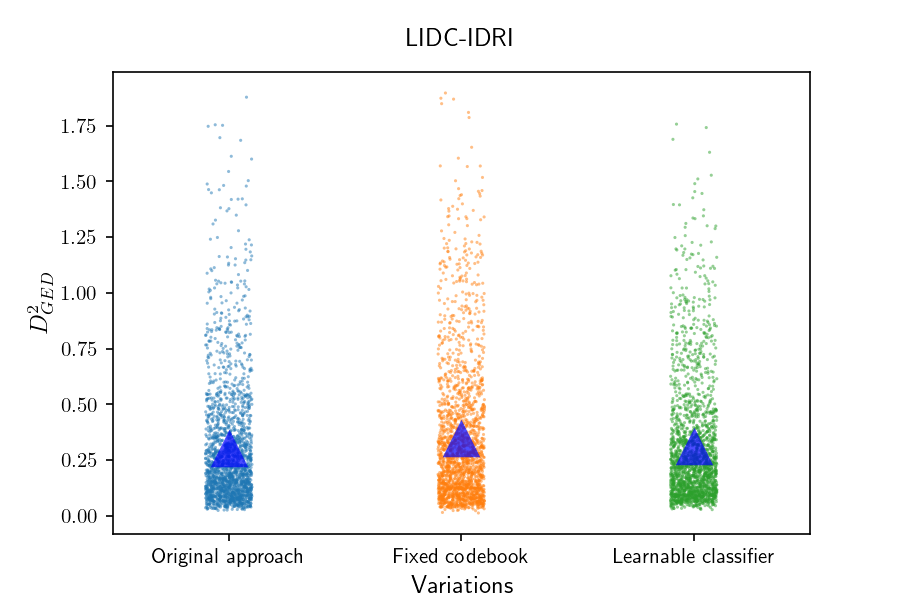}
    \includegraphics[width=2.8in]{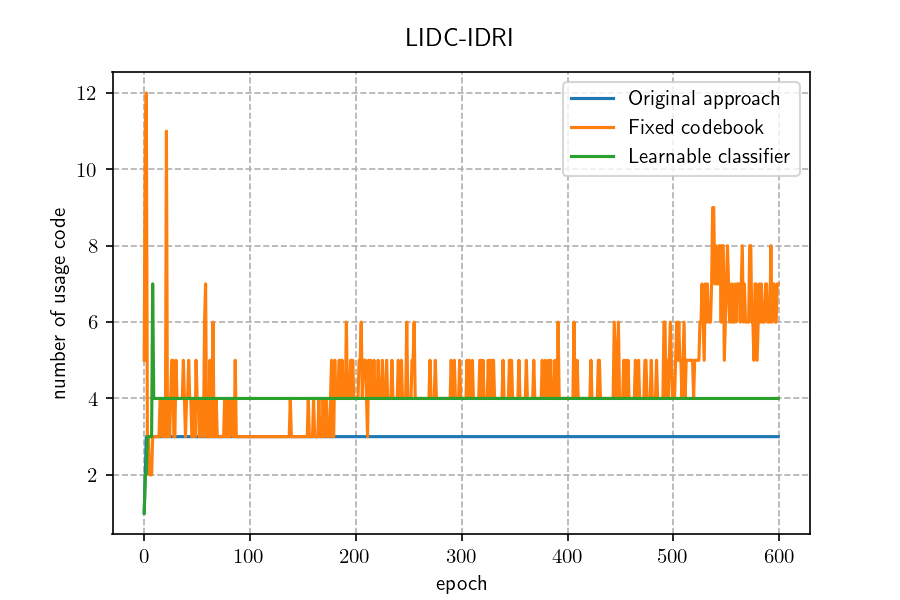}
	\caption{Ablations visualization. The left is the comparison of trick variations of our approach using the generalized energy distance. The small dots represent the quantities for test samples
    , and the small triangles represent the means of quantities. The right shows the number of latent codes used in each epoch during training.}
	\label{variation} 
\end{figure*}

In addition, we explore the performances of the renewal of the codebook and fixed ETF classifier by training our model on the LIDC dataset. We introduce two variations to our original approach: the Fixed codebook and the Learnable classifier. The Fixed codebook approach involves no longer updating the codes once the codebook space is initialized. On the other hand, the Learnable classifier replaces the fixed ETF classifier with a commonly used linear classifier that can be learned to predict probabilities. To evaluate the performance of these variations, we measure the $D_{\mathrm{GED}}^2$ metric on the test dataset and analyze the usage of codes during training. The results, as shown in Fig. \ref{variation}, indicate that our original approach outperforms the variation approaches. We observe that not updating the codebook results in worse $D_{\mathrm{GED}}^2$ scores and leads to unstable code utilization. This is because the covariance loss no longer functions concurrently when the codebook is fixed. Furthermore, Fig. \ref{variation} demonstrates the advantages of the fixed ETF classifier over the learnable linear classifier. In the absence of explicit data distribution for learning in the LIDC dataset, the fixed ETF classifier provided more accurate distribution predictions with fewer classes. In summary, our findings highlight the superiority of our original approach, which incorporates both the renewal of the codebook and the fixed ETF classifier. This approach yields better performance in terms of accurate distribution prediction and stable code utilization.

\section{Conclusion}\label{sec_conclusion}
We have adopted dynamic multi-valued mapping to describe the uncertainty in practice. It is a multi-valued mapping with a probability measure for each output. A data-driven optimization problem is further designed to solve dynamic multi-valued mapping. We proposed a general deep neural network framework to achieve this. Meanwhile, a codebook is introduced to explain the corresponding relationship between input and output in multi-value mapping. It also includes an effective evaluation of prediction probabilities to capture uncertainty in the dataset. Through extensive validation of synthetic and realistic tasks, we have demonstrated the superior performance of our method compared to state-of-the-art approaches.

\begin{acknowledgements}
L.M. Lui is supported by HKRGC GRF (Project ID: ).
\end{acknowledgements}

%
%

\bibliographystyle{spmpsci}      
\bibliography{citation.bib}   


\end{document}